\documentclass[proceedings]{ccn}
\ccndoi{10.32470/vp0ddtg}  

\title{Nothing from Something: Can a Language Model Discover 0?}

\author{Phoebe Zeng, Thomas L. Griffiths, and Brenden M. Lake \\
  Department of Computer Science, Princeton University \\
  \email{\{phoebe.zeng,tomg,brenden\}@princeton.edu}}
  
\begin{document}

\maketitle

\begin{abstract}
    AI systems based on artificial neural networks are being developed with aspirations of pushing the boundary of human mathematical knowledge. A key question for these systems is how much they can reach beyond their training data. Mathematical discovery requires a strong form of out of distribution generalization; the ability to hypothesize genuinely new -- and potentially logically more powerful -- mathematical structures. It has been hypothesized that language abilities support such generalizations in human cognition. In this work, we use simple arithmetic as a case study for examining how modern AI models could expand their mathematical horizons, evaluating whether these models can independently discover the concept of ``zero''. We show that (1) language models of a GPT-2 size are unable to perform this generalization at test time regardless of language pretraining, but (2) models can improve substantially after training on tens or hundreds of examples of zero. Additionally, we find that language pretraining reduces the number of required examples by approximately $50\%$, showing that language abilities can scaffold mathematical discovery in neural models.  
\end{abstract}

\section{Introduction}

The mathematical abilities of large language models have been a recent area of focus for the AI community. While math was at first seen as a setting where large language models struggled in comparison to symbolic models, post-trained so-called ``reasoning'' models have demonstrated impressive performance in the domain of abstract mathematics. Both OpenAI and DeepMind claim models that achieve Gold-medal level performance on the 2025 International Mathematical Olympiad (IMO) competition problems, an honor awarded on average to just $8\%$ of human participants \citep{lockhart_advanced_2025}. In the past few months, start-up companies Axiom and Harmonic announced that they had autonomously solved 8/12 problems on this year's Putnam exam (top $1\%$ of human participant scores) and contributed to solving Erd\H{o}s problem \#1026 \citep{tao_story_2025}. 

Due to this recent progress, mathematicians have begun to reckon with the idea of a world where proof-based mathematics is automated \citep{venkatesh, drosser_ai}. However, to the best of our knowledge, the aforementioned models have been trained on a large amount of mathematical data comparable in kind and complexity to their test problems \citep{lightman_lets_2023,trinh_solving_2024}. Thus, while indicating exciting progress, these benchmark results do not demonstrate an ability to go beyond the structures the models had already acquired during training. This would seem a prerequisite to extending human mathematics in a meaningful way. 

The question of if and when minds are capable of this kind of generalization has long been studied in cognitive science. The developmental psychologist Jean Piaget articulated a general theory of learning that is characterized by exactly this kind of leap from one formal system to another \citep{piaget_structuralism_1970}. As characterized by Jerry Fodor, this position holds that ``[I]f you [characterize the computational capacities of the organism] for several different time slices of the organism \dots, what you would get is a fundamentally different galaxy of constraints on the organism's concepts. \dots [T]he logic instantiated by the system of concepts at any $i^{th}$ stage is weaker than the logic instantiated by the $i + 1^{th}$ stage'' \citep{piattelli-palmarini_language_1981}. Fodor believed this kind of learning to be impossible. Formally, he argued there is no way to bridge the gap between the current universe of concepts and a universe that contains more powerful concepts; this difference in conceptual apparatus is exactly what defines them. If we cannot express the higher-order concepts in the lower-level vocabulary, how could we hypothesize them? 

In this work, we consider perhaps the simplest case of this kind of generalization: Can a model trained on positive single-digit arithmetic generalize to zero? The addition of zero to our number system has its origins in 1800 B.C., and has nice parallels in cognitive development \citep{lamb_ancient}. Cognitive scientists researching the way children learn the natural numbers hypothesize that language ability plays an important role \citep{spelke_what_2003,carey_origin_2009}. Specifically, they hypothesize that language allows children to articulate informal concepts beyond their current conceptual reach, which may evolve into formal concepts as children map their experience onto the symbols. 

In our experiments, we test the hypothesis that language pretraining scaffolds the ability of models trained on positive, single-digit arithmetic to generalize to zero. Specifically, we show that: 
\begin{enumerate}
    \item Neural network models based on the GPT-2 architecture trained on positive, single-digit arithmetic do not generalize successfully to zero, regardless of language pretraining.
    \item Neural network models with language pretraining require approximately $50\%$ less few-shot data to achieve the same test accuracy as neural network models without language pretraining, suggesting language ability does support this kind of generalization. 
    \item Zero does constitute a special case; neural network models do not have trouble generalizing to other digits when held-out of training in an analogous manner.
\end{enumerate}

\section{Background and related work}

\subsection{Math and large language models}
Just a couple of years ago, math was an area of difficulty for state-of-the-art language models. Hendrycks and colleagues highlighted proficiency in math as being notably resistant to model scaling \citep{hendrycks_measuring_2021}, and even ChatGPT (versions released in 2023) and GPT-4 were shown to perform well below the ``graduate level'' of knowledge implied by exam benchmark results \citep{frieder_mathematical_2023}. This led to the development of dedicated datasets and foundation models for math \citep{azerbayev_proofnet:_2023,azerbayev_llemma:_2024}. 

However, the big strides we have seen in the past few years can largely be attributed to the constellation of ``reasoning'' focused research paradigms and the availability of high quality, supervised datasets. At least in theory, it's clear how to supervise the solution to a math problem or proof; reinforcement learning and reinforcement learning-adjacent techniques are a natural fit for this data. OpenAI first demonstrated the impact of process supervision (supervising each step of the model's solution) on the previously cited challenging benchmark \citep{lightman_lets_2023}, and then (likely) built on this to release their first ``reasoning'' model \citep{openai_learning_reason}. The exact details of the model (``o1'') have not been publicly released. Google's first ``Olympiad-level'' AI system did not make as explicit use of reinforcement learning techniques, but cited the development of synthetic data as a major contributor to the success of the system \citep{trinh_solving_2024}. 

While the details of most of today's highest performing mathematical models are not published by industry labs, we can learn from the high-performing open-sourced theorem proving models, such as Goedel-Prover \citep{lin_goedel-prover-v2:_2025} and DeepSeekProver2  \citep{ren_deepseek-prover-v2:_2025}. This work reveals an impressive operation that revolves around scaling the (often synthetic) training data. As model abilities have increased, the benchmarks have become increasingly sophisticated, but so has the training data available to the models. This is because the models now produce and supervise training themselves. 

More evidence for the importance of training data is provided by the recent effort towards models for ``auto-formalization'', or the task of translating natural language mathematics into a formal language, such as Lean \citep{Lean}. Formalized mathematics can be automatically verified for correctness, and so makes up an ideal source of training data. Math Inc., a start-up whose mission is to create ``verified superintelligence via autoformalization'' recently formalized the Strong Prime Number Theorem, a long standing challenge in the formalization community \citep{gauss}. 

In this work, we seek to understand when and how models might be able to jump \textit{beyond} their training data at test time in order to solve a problem. This is especially relevant to systems where models contribute to their own training data. The question is also distinct from, but informed by the literature on compositional and algorithmic generalization in computational models. Prior work has contributed datasets for measuring compositional generalization, and shown that models may fail to generalize compositionally when the relationship between training and testing data is determined by systematic rules \citep{lake2018generalizationsystematicitycompositionalskills}. In an analogous fashion, we aim to contribute new benchmarks for our notion of conceptual generalization and show under what circumstances models may fail or succeed. 

\subsection{The cognitive science of number}
Our work is also motivated by theories about what allows the human mind to succeed in these settings. Particularly salient to our setting is the way children learn the meaning of number words. Many children are able to count to ten before they can reliably perform the ``give me N'' task, which requires the child to provide the experimenter with N objects. As they learn to perform this task, which entails understanding the meaning of the number word ``N'', children often learn the meaning of numbers one through four sequentially, but afterwards can generalize to greater numbers \citep{wynn_childrens_1990}. We take this as evidence that they have become ``cardinal-principle-knowers'', rather than simply ``four-knowers'' or ``five-knowers''. Instead of having disjoint representations for each number, we deduce they now have a representation of the natural numbers. 

These observations are a main motivating example for the theory in cognitive science that humans acquire new concepts through a process of ``bootstrapping'' \citep{carey_bootstrapping_2004}. This theory proposes that we use placeholder symbols as a bridge to creating and understanding new formal concepts. In the case of children learning the cardinal principle, number words begin as placeholders, but evolve into meaningful formal concepts as children interact with the world and use their experience to inform their understanding of the symbols. Such a theory for how humans are able to span conceptual discontinuities builds on work that suggests children and even babies have structured representations of number, but that are not powerful enough to generate concepts like the natural numbers \citep{spelke_what_2003,carey_origin_2009}. 

\begin{figure}[t]
  \centering
  \includegraphics[width=\columnwidth]{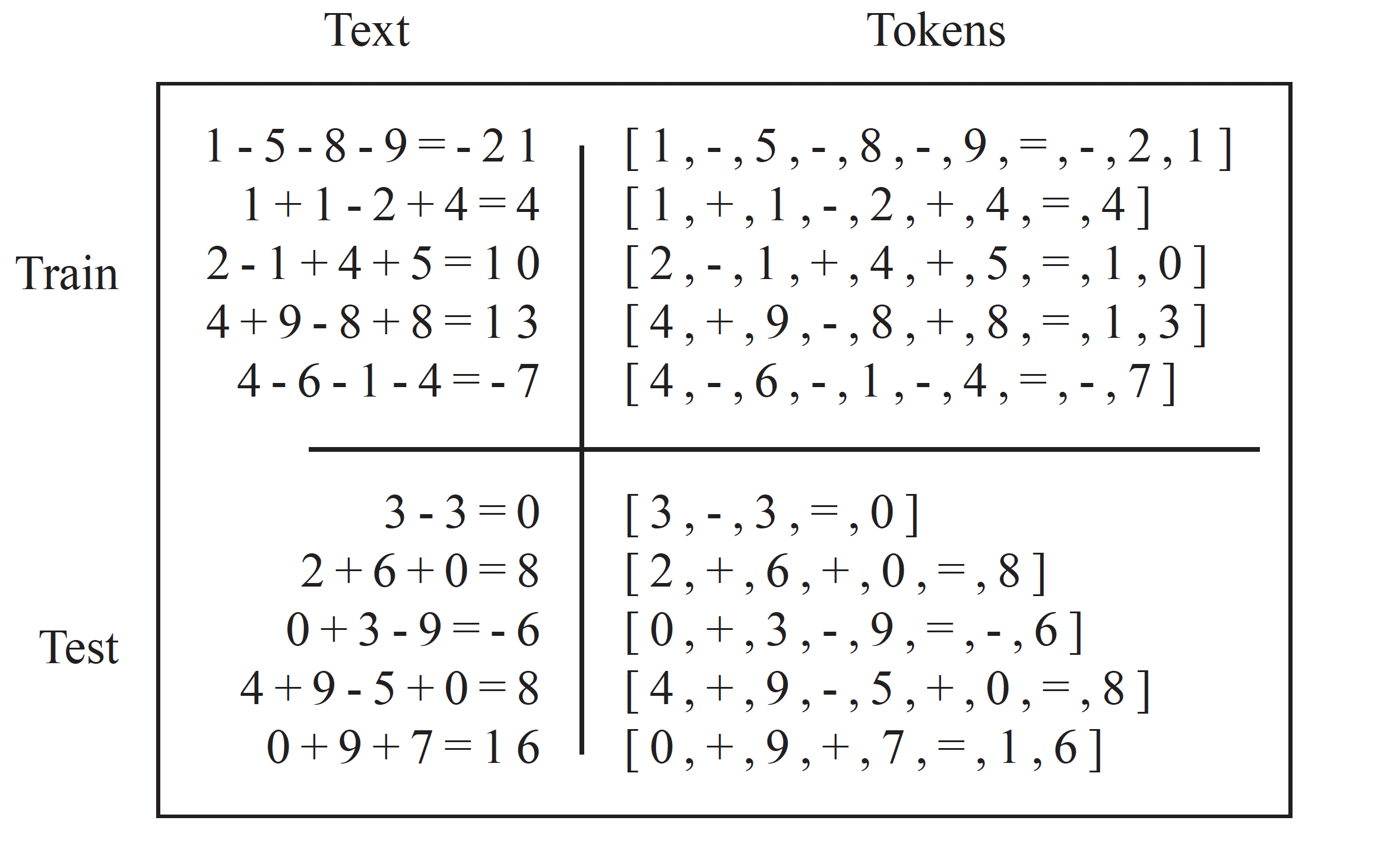}
  \caption{Example train and test data for arithmetic experiments. Examples that contain zero (except in the ones place) are held out and tokenization is per-digit. }
  \label{fig:data}
\end{figure}

\begin{figure*}[t]
  \centering
  \includegraphics[width=\textwidth]{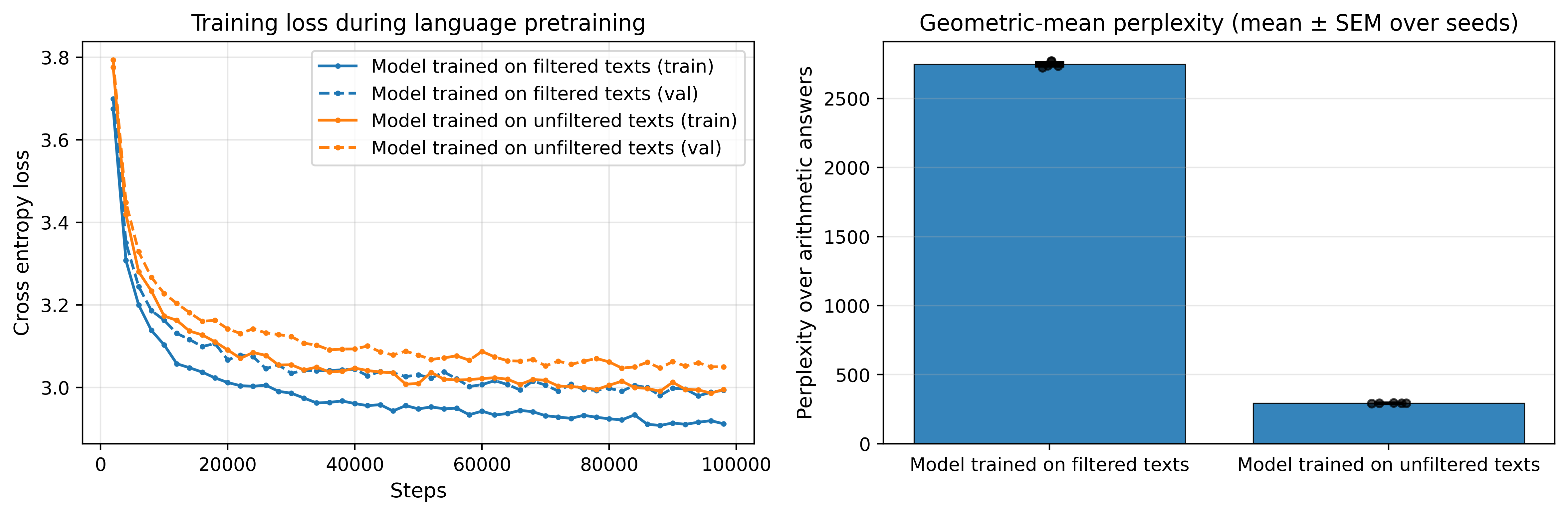}
  \caption{Language pretraining curves and model perplexity on arithmetic train data after language pretraining. The removal of text containing numbers results in a far greater perplexity for arithmetic problems, confirming that the model has little exposure to the relevant concepts prior to the training we provide on arithmetic.}
  \label{fig:pretrain}
\end{figure*}

The number zero, in particular, may be a challenging concept to learn. The origin of zero can be traced back to the Babylonians, who though they did not understand zero as a number, did use a symbol to indicate the difference between ``3601'' (written 101 in base 60) and ``61'' (written 11 in base 60) \citep{lamb_ancient}. It took over 240 decades to arrive at a formal definition of zero as a number, as Brahmagupta details in 628 A.D. \citep{kaplan_nothing_2000}. It has been suggested that the influential concept of ``nirvana'' may explain why Indian culture gave rise to this mathematical invention \citep{bellos_nirvana_2013}. Across cultures, the introduction of zero was anything but smooth \citep{seife_zero:_2000}. In this work we examine if the circumstances and features -- be they cognitive or societal -- that gave rise to the invention of zero in human mathematics can be replicated artificially. 

\section{Methods}

We consider the simple setting of arithmetic in our experiments. We train models on single-digit arithmetic problems of variable length that do not contain zero, and evaluate their ability to generalize to arithmetic problems that do include zero at test time. We consider models with and without language pretraining, controlling for mathematical knowledge implicit in the language data. In the following sections we describe the models, data and tasks in more detail, as well as our experimental design. 

\subsection{Models}

All models we consider are GPT-style decoder-only transformers \citep{radford2019language}; we vary only the number of layers, attention heads, and embedding dimension. In particular, we consider two configurations of these parameters, a small model with 4 layers, 4 attention heads, and an embedding dimension of 128 (800K trainable parameters) and a larger model with 12 layers, 12 attention heads, and an embedding dimension of 768 (124M trainable parameters). Throughout our experiments, we test various subsets of (1) the GPT-2 sized model pretrained on the unfiltered, truncated OpenWebText corpus (2) the GPT-2 sized model pretrained on the filtered OpenWebText corpus (3) the GPT-2 sized model without any pretraining, and (4) the smaller transformer without any pretraining. Lastly, we include some larger open-sourced models that are pretrained on data that includes numbers and math in the Appendix, as an upper bound of how well we might expect language models to pick up ``zero''.

\subsection{Datasets}

\subsubsection{Arithmetic}

In order to generate diverse examples with a final result of zero, we consider both addition and subtraction operations on the domain of single digit integers (zero to nine). Problems are constructed to have two to four single-digit operands on the left hand side (LHS), and a single result on the right hand side (RHS). We randomly sample all components of the LHS: (1) the number of terms (2) the terms themselves and (3) the operations between terms, and deterministically compute the RHS. For example, we first sample a number $n$ between two and four (inclusive), then sample $n$ digits and $n-1$ operations, and finally construct the data point (e.g. ``3+4-2=5''). Sequences are tokenized manually in order to ensure that each digit and symbol is encoded using its individual token. Importantly, this means two digit answers such as ``10'' or ``20'' are represented to the model as a ``1'' (or ``2'') followed by a ``0'', rather than as a distinct token. Examples of the data, pre and post-tokenization can be seen in Figure \ref{fig:data}. 

This manual tokenization scheme differs from the tokenizers used to pretrain our language models. However, as is detailed in the following section, our main model of interest is trained on a filtered corpus that does not contain any instances of arithmetic or mathematical symbols. So these two different tokenization schemes do not constitute a conflict between what the model sees during language pretraining and arithmetic training. This is not true of the control model we train on the unfiltered corpus; this should be noted as a potential reason the unfiltered model does not further outperform the filtered model. 

Previous work showed that when training Llama-2 (7B) to learn arithmetic, reversing the answer of the arithmetic problems improved model performance (i.e., an answer ``15'' gets encoded as ``51'') \citep{shen_revorder:_2024}. It is hypothesized that this makes it easier for the model to learn the ``carry'' algorithm. However in our setting, we felt reversing the answer to arithmetic problems could limit the value of language pretraining. Since the previous research is not directly transferrable to our setting due to model size and data representation, we do not reverse the answers in our arithmetic problems. For completeness, we do report results for many of our experiments with the answer reversed in the Appendix. 

Lastly, since the problems are of variable length, we append an ``End-of-Sequence'' (EOS) token to the end of the sequences so that the model learns when to finish. In order to generate our final train, validation and test sets we generate problems randomly as described above, and allocate any problems where zero appears on the LHS or is exactly equal to the RHS to the test set. We include problems where zero appears in the ones digit of the answer (e.g. ``5+5=10'') in the train set. This ensures, along with manual tokenization, that the token corresponding to zero accounts for 2.5\% of supervised tokens seen during training. We generate $10,000$ training samples, and $1,000$ samples each for the validation and test sets. \\

\begin{figure*}[t]
  \centering
  \includegraphics[width=\textwidth]{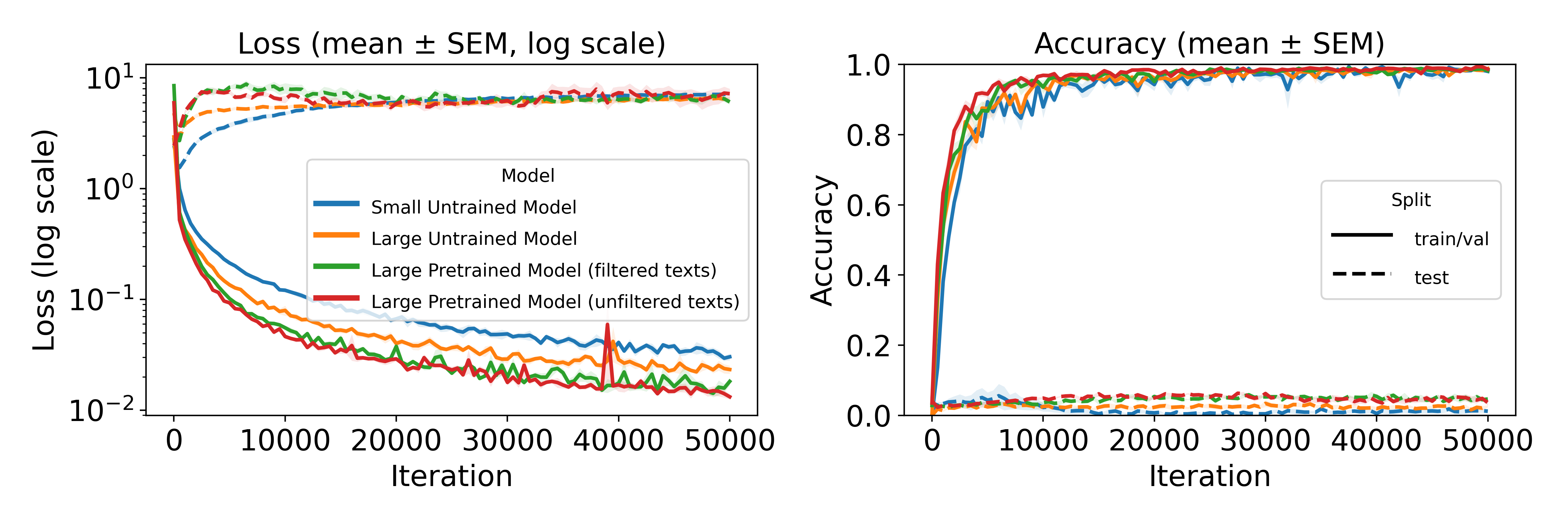}
  \caption{Comparison of model generalization to zero at test time, across training regimes. The training and validation sets consist of arithmetic problems that do not contain zero, while the test set consists entirely of arithmetic problems that do. All models were trained on the same data, but varied in size and whether they were pretrained on text that was filtered to remove mentions of numbers or arithmetic symbols.}
  \label{fig:test-time}
\end{figure*}

\subsubsection{OpenWebText}

Determining what a language model may or may not have seen during training is a frequently cited limitation to experiments involving pretrained language models, even when the models are open-sourced. In order to guarantee that our language model is not exposed to any math during training, we curated our own version of OpenWebText, a popular open-sourced corpus for pretraining English language models. 

OpenWebText is an open-source recreation of GPT-2's training corpus, based on the methods described in their original paper \citep{radford2019language}. The corpus is sourced entirely from Reddit posts. More specifically, the corpus is sourced from the Reddit submissions dataset which contains $8,013,769$ documents \citep{Gokaslan2019OpenWeb}. The creators of OpenWebText then de-duplicated the posts, and filtered out entries that were not English or valid HTML.  

For this work, we further filtered this corpus by removing any content that did not consist entirely of natural-language (no numbers, symbols, etc.), and more specifically English. The most naive methods to filter this content would have resulted in a dataset a small fraction of the size of the original. In order to maintain text volume, we performed the following cleaning steps. We first removed HTML artifacts, emails and footnotes from the texts, and then filtered out any texts that could not confidently be identified as English. We then converted all numbers less than one million to number words, and attempt to convert common trailing units (e.g., ``100\%'' is converted to ``one hundred percent'' and ``50\$'' is converted to ``fifty dollars''). We do the same for ordinals (e.g., ``1st'' is converted to ``first''). Excepting dates and times in their most common formats (we left these in), we then applied our natural language filter, removing texts that contained any non-natural language characters, including mathematical symbols like ``+'' or ``=''. This pipeline left us with 45.1\% of the documents, which we used in their cleaned form for training. We tokenize the texts using GPT-2's open-source tokenizer \citep{radford2019language}.

As further confirmation that we had filtered out the relevant texts, we also pretrain on a truncated version of the original dataset (matched to the size of the filtered dataset). We measure both models' perplexity on five different randomly sampled arithmetic train datasets after language pretraining, and show that the model trained on the unfiltered corpus scores significantly better than the model trained on the filtered corpus. The perplexity scores of the model trained on the filtered corpus are consistent with models unfamiliar with these tokens. These results are visualized in Figure \ref{fig:pretrain}.

\subsection{Language Modeling}

The models we pretrain on language are of the same size and architecture as GPT-2. As mentioned above, they have 12 layers, 12 attention heads, and an embedding dimension of 768 (124M parameters). We train one model on the filtered OpenWebText corpus, and another on the unfiltered truncated OpenWebText corpus. Both datasets are described in detail above. Since language model training is not the primary focus of the project, we make standard choices whenever possible. Tokenized sequences are batched into inputs of a fixed length (1024 tokens), and the remaining left-over tokens from each document are not kept or padded. We use cross-entropy loss as our training objective, and loss is computed for all tokens in the sequence. We minimize our objective using the AdamW optimizer with standard parameters ($\beta_1=0.9, \beta_2 =0.95$, weight decay $=0.1$). Weight decay is only applied to 2-dimensional parameters; biases and LayerNorm parameters are not decayed. We use a cosine learning-rate scheduler, with warm-up. 

The models are trained for $98,000$ steps, where a step consists of a batch of $480$ samples of length $1024$ tokens. After training, the model trained on the unfiltered data reaches a train loss of $2.99$, while the model trained on filtered data reaches a train loss of $2.9$. Training convergence of both models can be seen in Figure \ref{fig:pretrain}. We used two GPUs to train each model; each training run took approximately $48$ hours. 

\subsection{Arithmetic Training}

During arithmetic training, all sequences are tokenized manually, as motivated in the dataset description. However, the pretrained models are trained on sequences of tokens as indexed by GPT-2's tokenizer, while other models have a minimal vocabulary comprised of the digits 0-9, the requisite arithmetic symbols, and EOS and PAD tokens. The sequences are right-hand padded, and loss is only computed on the RHS (the answer to the arithmetic problem), up to and including the EOS token. Neither the LHS (including the ``='') nor the padding tokens contribute to loss. Accuracy is computed by providing the model with the LHS (including the ``='') as context, generating an answer using greedy decoding and checking if that answer is correct. The longest correct answer is at most four tokens long.

Training hyper-parameters are held constant across models. We use a constant learning rate of $0.0001$, a batch size of $64$, and minimize our objective (cross-entropy loss) using the AdamW optimizer with standard parameters ($\beta_1=0.9, \beta_2 =0.95,$ weight decay $=0.1$). Weight decay is only applied to 2-dimensional parameters; biases and LayerNorm parameters are not decayed. We use dropout of $0.1$ on model activations for all experiments except those that measure generalization to other numbers. In this case, we turn off dropout to isolate systematic differences across digits (not due to stochasticity) since we only care about relative performance. In order to be compatible with our pretrained language models, we do not report results that train a bias in Linear and LayerNorm parameters. 

\section{Experiments}

\subsection{Zero-shot generalization}

Our first experiments test whether or not the models can generalize to arithmetic problems that contain zero at test time, after training only on arithmetic problems that do not contain zero. As detailed in the sections above, the models are exposed to the zero token during training only when it appears in the ones-place of the answer.

We train each model for $50,000$ steps and evaluate accuracy and test loss every $100$ steps. The model sees one batch of data ($64$ data points) each step. The validation set, like the training set, does not have any examples containing zero, and is only used to evaluate model accuracy. We compute both loss and accuracy metrics on the test set. We report average metrics across five random seeds, computing the standard error of the mean. 

\subsubsection{Results}

Loss and accuracy metrics over the entire training course are visualized in Figure \ref{fig:test-time}. The results for the test-time generalization experiments are clear: none of the models we test demonstrate the ability to generalize to zero. Training loss trajectories are also cleanly separable, the GPT-2 sized model pretrained on unfiltered, truncated OpenWebText achieves the lowest training loss throughout training, followed by the GPT-2 sized model pretrained on filtered OpenWebText, the untrained the GPT-2 sized model, and finally the small transformer model. Nonetheless, as training loss decreases and validation accuracy saturates, test loss and accuracy show no signs of improving, across models.

\subsection{Few-shot learning}

Our second set of experiments measures how well the models are able to generalize at test-time when we allow a small fraction of the training data to include zero. Put simply, the second set of experiments is identical to the first, except we allow a specified number of ``few-shot learning'' examples to be mixed into the training dataset. We vary the number of training samples that contain zero from $2^0=1$ to $2^{10}=1024$.

We run this set of experiments only for (1) the GPT-2 sized model pretrained on the filtered OpenWebText corpus (2) the GPT-2 sized model without any pretraining. Since we are mainly interested in relative accuracies rather than absolute values, we train each model for $25,000$ steps, rather than $50,000$. All other parameters are unchanged from the first set of experiments. 

\begin{figure}[t]
  \centering
  \includegraphics[width=\columnwidth]{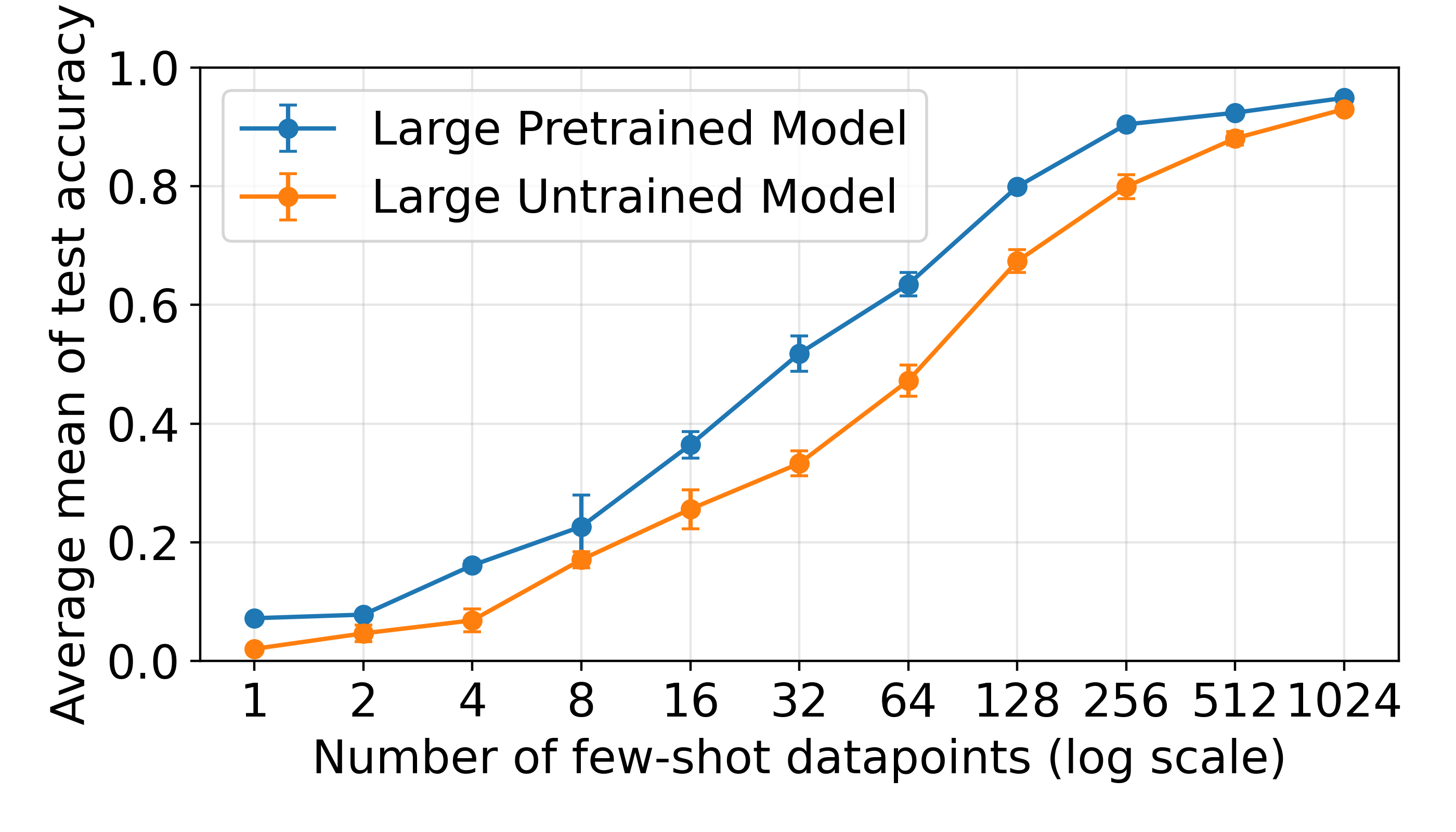}
  \caption{Model generalization to zero at test time in few-shot regime. Test accuracy is averaged over last 1000 steps of each training run; mean and standard error are reported over five random seeds.
  Accuracy on problems involving zero increases rapidly with more examples, and faster for the model with language pretraining.}
  \label{fig:few-shot}
\end{figure}

\subsubsection{Results}
The average test accuracies across our different numbers of few-shot data points are visualized in Figure \ref{fig:few-shot}. The results for the few-shot learning experiments reveal that while models struggle under the first training regime, a small amount of few-shot examples makes a meaningful impact on the model's ability to generalize at test time. 

For the GPT-2 sized model pretrained on filtered OpenWebText, just 64 examples (0.64\% of the training data) is enough to achieve a test accuracy above 60\%, and by 1024 examples (10.24\% of the training data) the model robustly achieves an average test accuracy greater than 90\%. 

The untrained GPT-2 sized model also benefits dramatically from the few shot examples introduced to the training set, but consistently requires more data than the model that has benefited from language pretraining. 

To quantify the sample-efficiency benefit of language pretraining, we estimate how many few-shot examples an untrained model would need to match the accuracy achieved by a pretrained model. For each seed and each few-shot budget $N$ on the pretrained curve, we take the pretrained test accuracy as the target and use log-scale linear interpolation between measured few-shot budgets to estimate the total number of few-shot examples the untrained model requires to match that accuracy. We then define percent data reduction as $1 - N_{\text{pretrained}} / N_{\text{matched untrained}}$. For each seed, we average reduction over all few-shot budgets; across seeds, the mean reduction is $48.5\%$ (bootstrap $95\%$ confidence interval: [$41.2\%$, $55.9\%$]; $n = 5$ seeds). A one-sided one-sample t-test on these per-seed means ($H_{0}$: mean reduction $\leq 0$) yields $p = 1.7 \times 10^{-4}$.

\begin{figure}[t]
  \centering
  \includegraphics[width=\columnwidth]{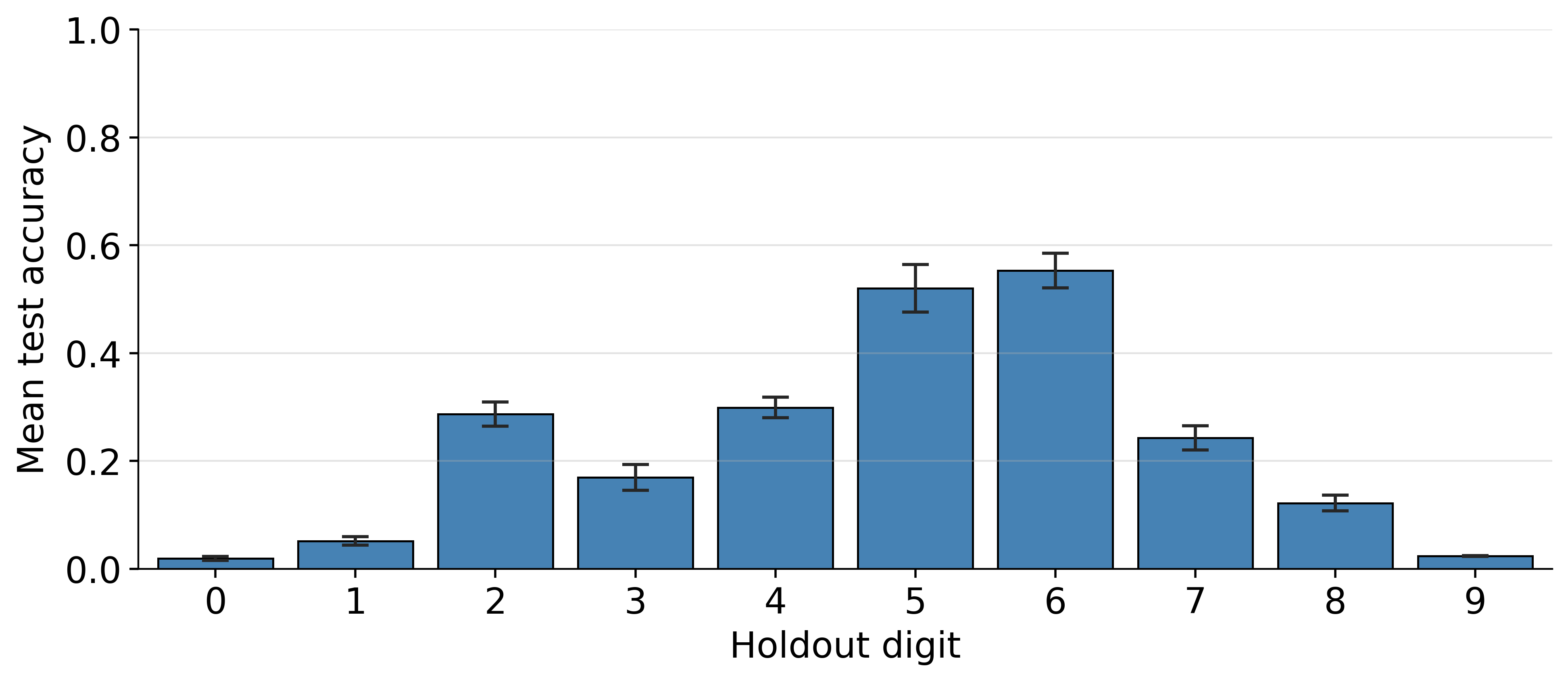}
  \caption{Final test accuracy on holdout digits 0-9. Zero and nine (carry digit) are hardest to generalize to at test time.}
  \label{fig:zero-special}
\end{figure}

\subsection{Other numbers}

Our last set of experiments is designed to answer the question: Is zero special? In these experiments, we run the same tests for zero-shot generalization, but instead of holding out zero for the test set, we remove the other digits (one to nine), each in turn. Specifically, for each digit one through nine, we generate train, validation and test datasets using an identical procedure as that described for zero. We train the model for $25,000$ steps in this set of experiments, and leave all other parameters unchanged from the first two sets of experiments. 

In order to study how proximity to the digit that initiates the ``carry'' operation impacts test-time generalization, we also perform these experiments in the base-8 setting. In this case, we examine the ability to generalize to digits (zero to seven) at test-time. All the experiments to test generalization for other numbers are run only using the GPT-2 sized model pretrained on the filtered OpenWebText corpus.

\subsubsection{Results}
The results for our last set of experiments are visualized in Figures \ref{fig:zero-special} and \ref{fig:base8}. Full loss and accuracy trajectories are given in the Appendix; in these plots we focus on the final test accuracy achieved for each digit. 

In both the base-10 and base-8 cases, digits that fall in the middle of the range are easier to generalize to at test time, with difficulty increasing towards both zero and the carry digit. Given the additional context of the parallel results for holding out seven in the base-8 case, we hypothesize that the difficulty with holding out nine in the base-10 case can be attributed to this digit's special status of being the digit that initiates the carry operation. We attribute the increased difficulty of generalizing to one to the fact that it appears in the tens column of the results a majority of the time. However, the pattern is stronger than these individual explanations. 

One hypothesis is that this result can be explained by the model's ability to interpolate and inability to extrapolate. This hypothesis holds that digits like 4 or 5 are easier to interpolate because the model can make use of more neighboring digit representations than are available for digits such as 0 or 9. If this hypothesis were true, we would expect to see that (1) the model represents the held-out digit similarly to its neighbors, and (2) there are more similar neighbors for digits like 4 and 5 than there are for 0 or 9. In order to provide some initial evidence for this hypothesis, we compute the cosine similarity of the model's representation (embedding vector) of the held-out digit to all other digits. We then plot in Figures 7 and 8 the number of these digits that have a cosine similarity of 0.65 or greater to the held-out digit's representation. We see that indeed the plot is roughly an inverted-V shape. Representations of digits 0 and 9 have high similarity with fewer digits than do digits 4 and 5. 

\begin{figure}[t]
  \centering
  \includegraphics[width=\columnwidth]{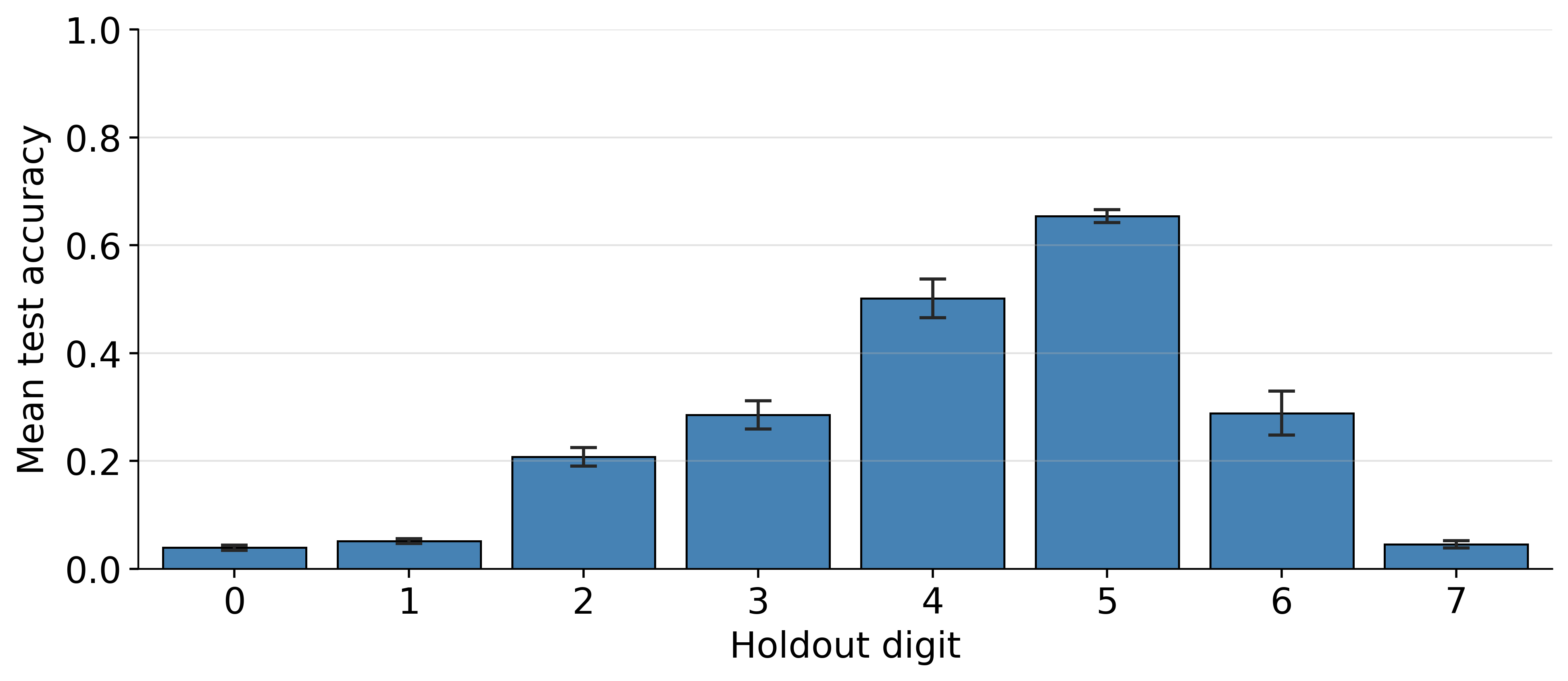}
  \caption{Final test accuracy on holdout digits 0-7, in the base-8 arithmetic regime. Zero and seven (carry digit) are hardest to generalize to at test time.}
  \label{fig:base8}
\end{figure}

\section{Limitations and future directions}

Understanding whether systems based on large language models have the potential to create new mathematics is a challenge for both artificial intelligence researchers and cognitive scientists. In this paper, we have explored perhaps the simplest version of this problem, investigating whether language models (based on the GPT-2 architecture) trained on arithmetic in a world without zero can postulate it when it is required to solve a new set of problems. Based on research in cognitive science, we hypothesized that training on language in general might facilitate this form of generalization. Overall, our results support two qualitative conclusions: (1) Generalizing to new mathematical concepts is non-trivial even in simple cases for models like GPT-2. (2) Facility with language does support this kind of generalization. These conclusions raise many new questions: Why are some digits harder to generalize to than others? How do representations learned during language pretraining support generalizing to these unfamiliar tokens? In this section we identify some of the limitations of our work and highlight a few salient next steps in this line of research. 

\subsection{Transfer of language representations}

Our results showed that pretraining on language can make models faster to learn new mathematical concepts. This raises interesting questions about what role language is playing in this process. What has the language model learned during pretraining that allows it to generalize faster to zero? Is the model able to transfer learned linguistic concepts like ``nothing'' (or similar) to the arithmetic setting? Can we observe specific circuitry within the network responsible for these generalizations? While not explored in this work, this is an important line of questioning for follow-up work. 

\subsection{Training techniques}

The class of models we focused on in this paper, based on the GPT-2 architecture, provides a way to explore how training on language influences generalization. However, more recent work with larger models has highlighted other ways in which these models can be prompted and trained. In particular, prompting and training techniques focused on ``reasoning'' have become very popular \citep{li202512surveyreasoning}. Training language models to generate intermediate work (as if using a scratch-pad) before reporting the final answer has been show to dramatically improve performance on tasks that involve many steps, including arithmetic \citep{nye_show_2021}. Relatedly, prompting language models to generate intermediate reasoning traces has been shown to be similarly effective on diverse reasoning problems \citep{wei_chain--thought_2023}. It would be interesting to explore whether or not these techniques aid the models ability to generalize to unseen tokens like zero at test time. 

\begin{figure}[t]
  \centering
  \includegraphics[width=\columnwidth]{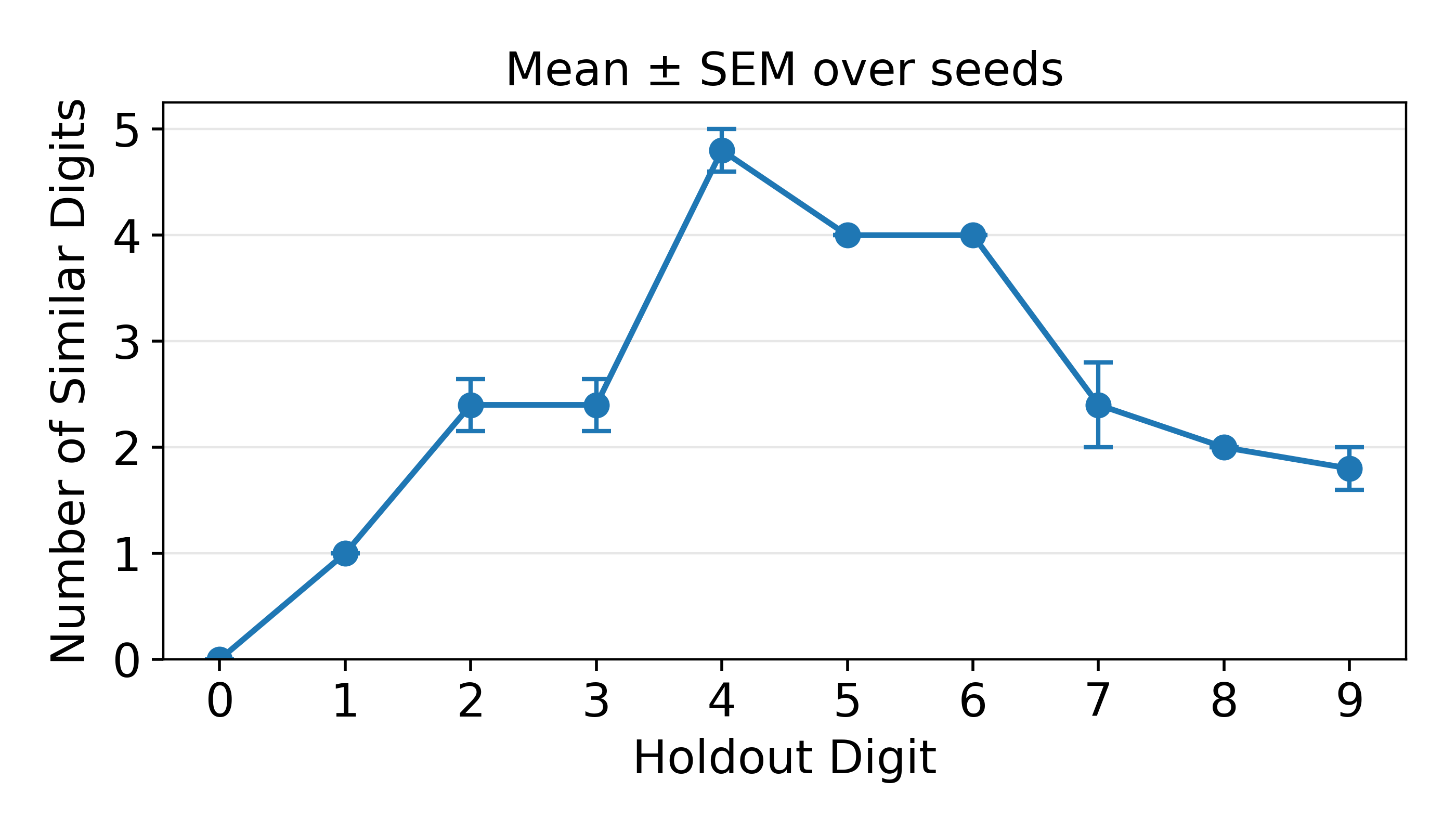}
  \caption{Number of digits with cosine similarity $\geq$ 0.65 with holdout digits 0-9. Digits that fall in the middle of the range have more ``near neighbors''.}
\end{figure}

\subsection{Model and data scale}

Simply scaling the size of language models and their datasets, in combination with longer training has yielded dividends for AI researchers. So-called ``scaling laws'' for language models have been studied and debated as such \citep{kaplan_scaling_2020}. The interaction of model scale with desirable attributes like few-shot / in-context learning has also been shown \citep{brown_fewshot_2020}. In the Appendix, we test 2 1-billion parameter open sourced models, and show they too cannot discover zero at test time. In future work we plan to test models of increasing scale in order to understand how model and data scale interact with the generalization abilities we study in this paper. The size of our models, datasets, and training time may all contribute to the negative result we report in our first set of experiments. This kind of scaling up is also likely to be important to access the full potential of techniques like reasoning and chain-of-thought prompting.

\begin{figure}[t]
  \centering
  \includegraphics[width=\columnwidth]{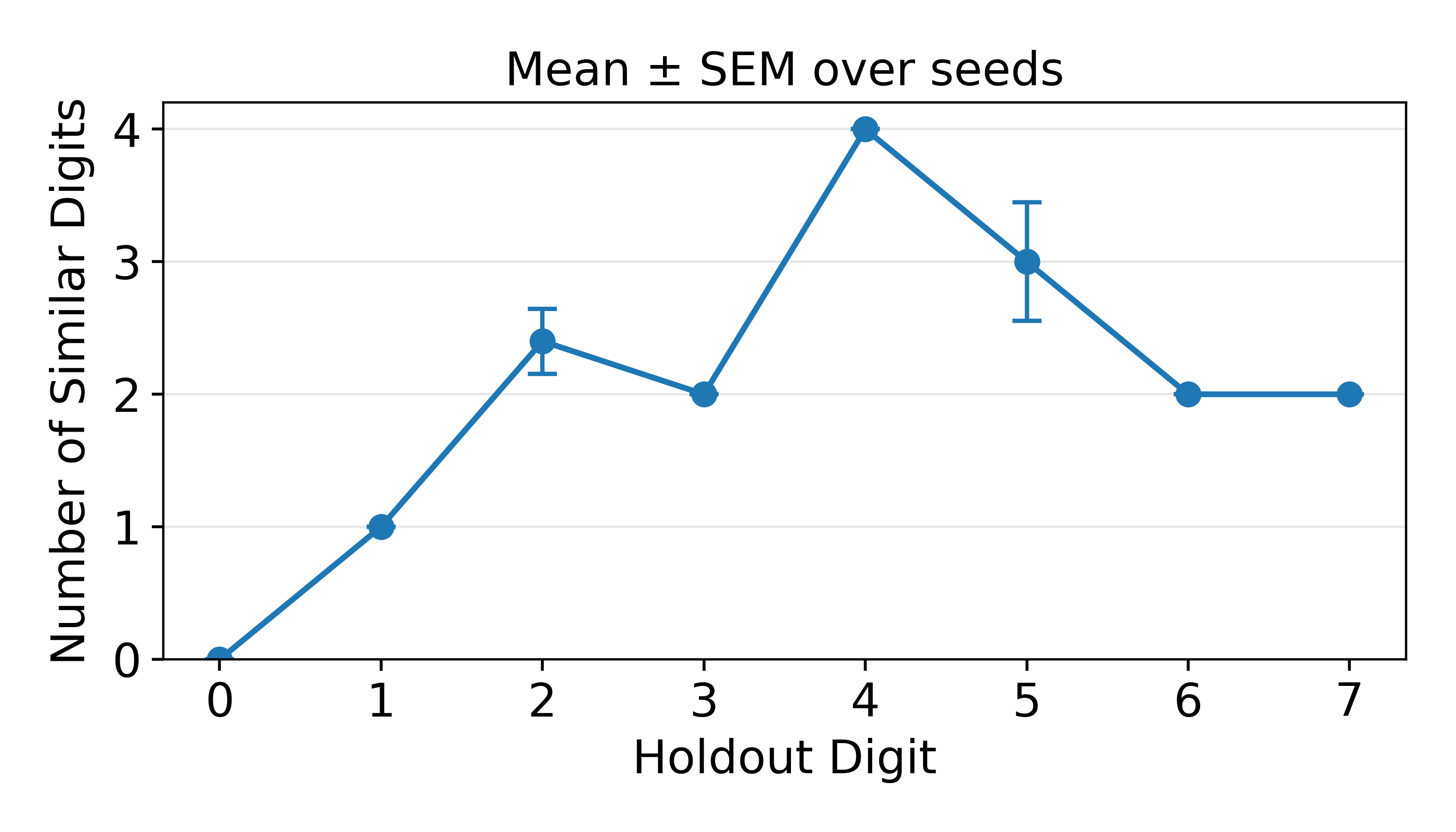}
  \caption{Number of digits with cosine similarity $\geq 0.65$ with holdout digits 0-7, in the base 8 arithmetic regime. Digits that fall in the middle of the range have more ``near neighbors''.}
\end{figure}

\section{Conclusion}

Mathematical benchmarks are saturating quickly. Does this suggest that today's models are close to contributing meaningful new advances in the field? In this work we show that even simple arithmetic is a challenging setting for measuring the ability of language models to expand their mathematical boundaries. In particular, we show that models with the architecture and size of GPT-2 struggle to understand zero when it's not included in the training data. This remains true even when the model has received language pretraining. However, reminiscent of the cognitive science theory that language scaffolds conceptual evolution in human cognition, we do find evidence that language pretraining supports this kind of test-time generalization in neural models. In particular, we show that models with language pretraining require approximately 50\% less few-shot data to achieve the same test accuracy as models without language pretraining. Overall, our experiments suggest that if we hope to develop models that can truly contribute new mathematical results, we have much to learn from simple settings where such generalization can be rigorously measured. 

\section{Acknowledgments and Disclosure}

The use of AI in the work was restricted to code generation. Model families including GPT, Gemini, and Claude models were used for this purpose. 

\printbibliography

\section{Appendix}

\subsection{Results on open-sourced models}

We also consider two open-sourced pretrained language models: the 1-billion parameter versions of the Llama-3.2 \citep{grattafiori2024llama3herdmodels} and Pythia \citep{biderman2023pythiasuiteanalyzinglarge} series of models. Both models are 16-layer decoder-only transformers with an embedding dimension of 2048. Llama-3.2-1B is distilled from larger Llama-3.1 models, and trained on up to 9T tokens of public web text and code. Pythia-1B is trained on $\approx$300B tokens from The Pile \citep{gao2020pile800gbdatasetdiverse}.

Since these models have seen arithmetic during pretraining, we define new tokens for all those required to encode arithmetic problems. I.e., since the model saw the token ``0'' during its pretraining, we extend the model's vocabulary to include a new ``0'' token that we use to encode our arithmetic data. We do this for all tokens present in the arithmetic data (for all digits, + -, =). The new token embeddings are initialized via sampling from a multivariate normal whose mean is the mean of existing embeddings and whose covariance matches the empirical covariance of existing embeddings \citep{Hewitt}.

Using this paradigm, we report results on these models for the zero-shot experiments. We use a learning rate of $0.00003$ and inherit dropout and bias configurations from the pretrained configurations. Optimizer settings and those not mentioned by name here are matched to the zero-shot experiment described in the main text. 

Figures 9 and 10 show loss and accuracy trajectories during training. These models do not demonstrate any improved ability to generalize to zero in comparison to the smaller models we tested.

\begin{center}
\includegraphics[width=0.5\textwidth]{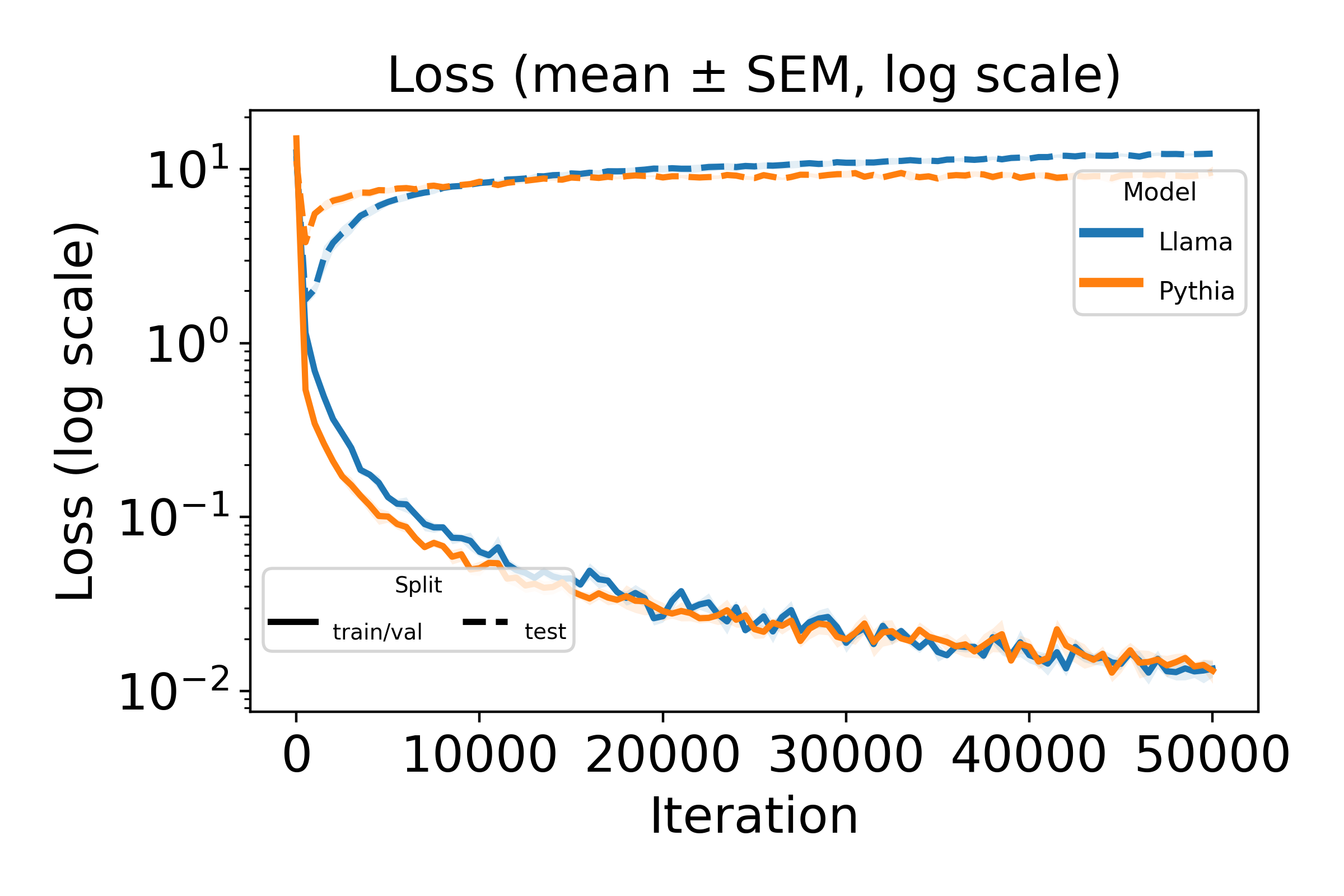}
\end{center}
\noindent \small{Figure 9: (Open sourced) model generalization to zero at test time.}

\begin{center}
\includegraphics[width=0.5\textwidth]{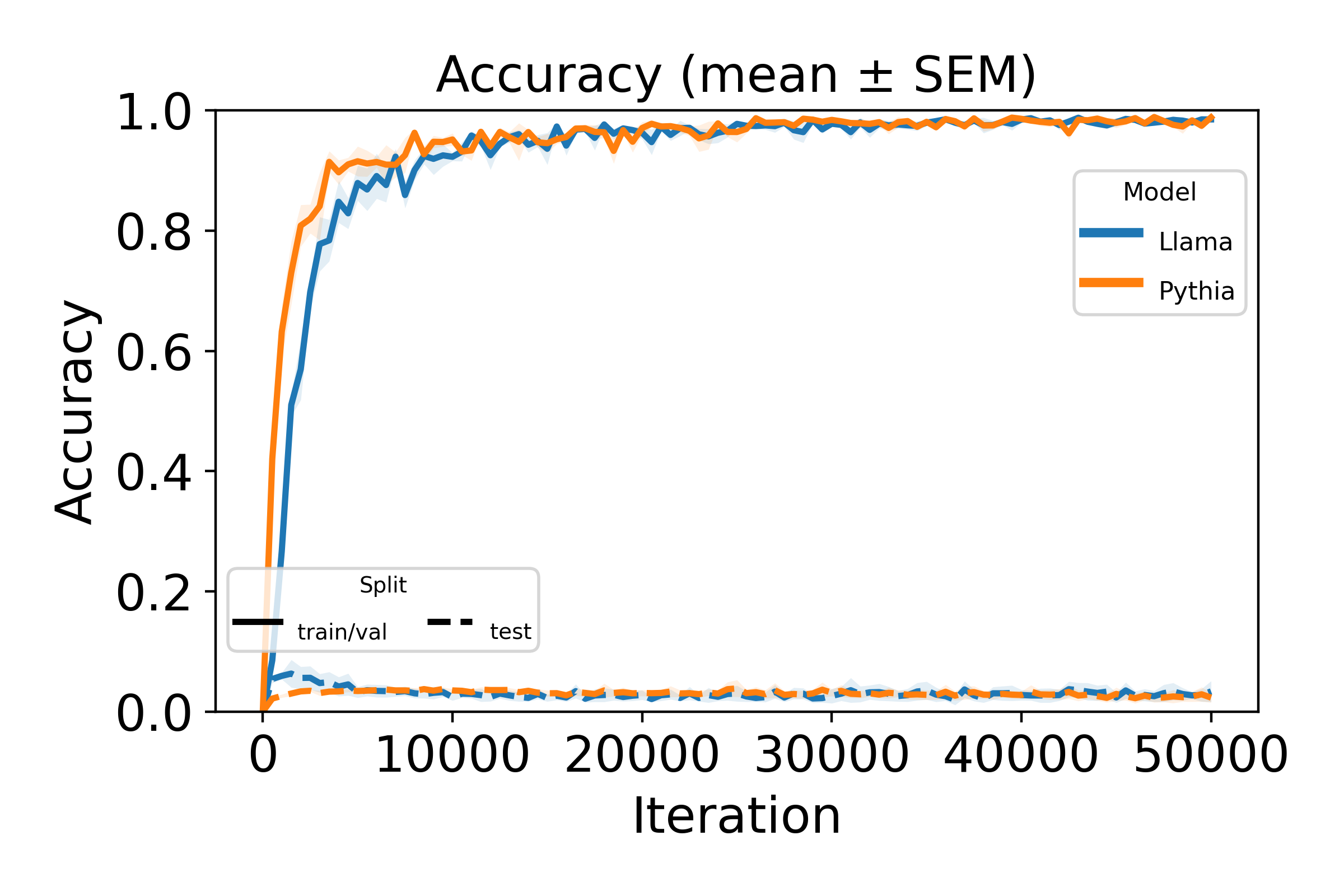}
\end{center}
\noindent \small{Figure 10: (Open sourced) model generalization to zero at test time.}

\subsection{Results with answer reversed}
In order to address uncertainty surrounding the effect of reversing the answers to the arithmetic problems we present to our models, we repeat many experiments with reversed answers. All other settings remain constant from those we report in the main text.

\subsubsection{Zero-shot generalization}

While we originally hypothesized that reversing the answer would limit the advantage of models that saw arithmetic during pretraining, we see a slight increase in the performance gap between the model trained on filtered texts and the model trained on unfiltered texts with the answer reversed. Figures 11 and 12 show loss and accuracy trajectories during training for the main models, and Figures 13 and 14 show loss and accuracy trajectories during training for the open sourced models.

\begin{center}
\includegraphics[width=0.5\textwidth]{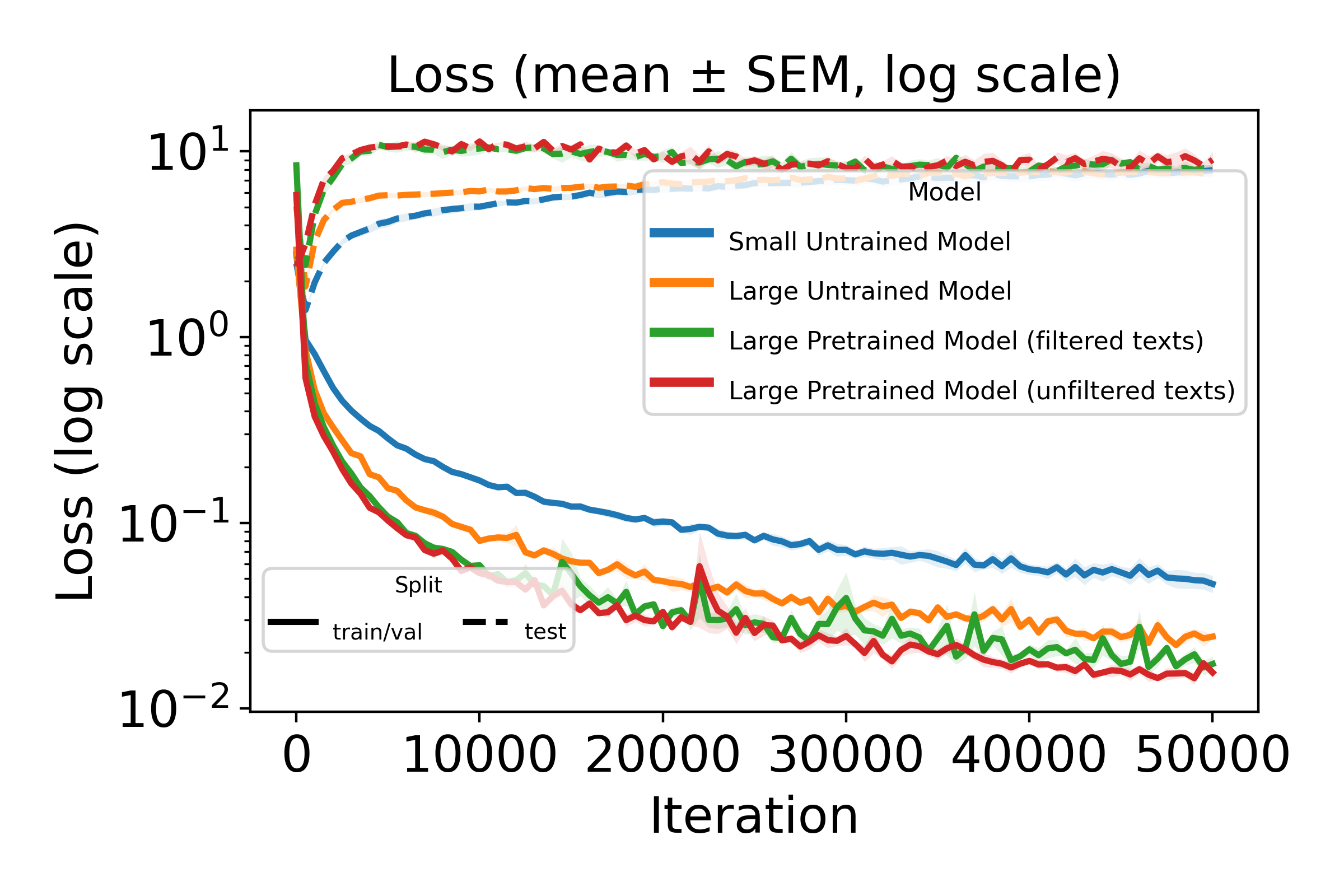}
\end{center}
\noindent \small{Figure 11: Model generalization to zero at test time with answer reversed.}

\begin{center}
\includegraphics[width=0.5\textwidth]{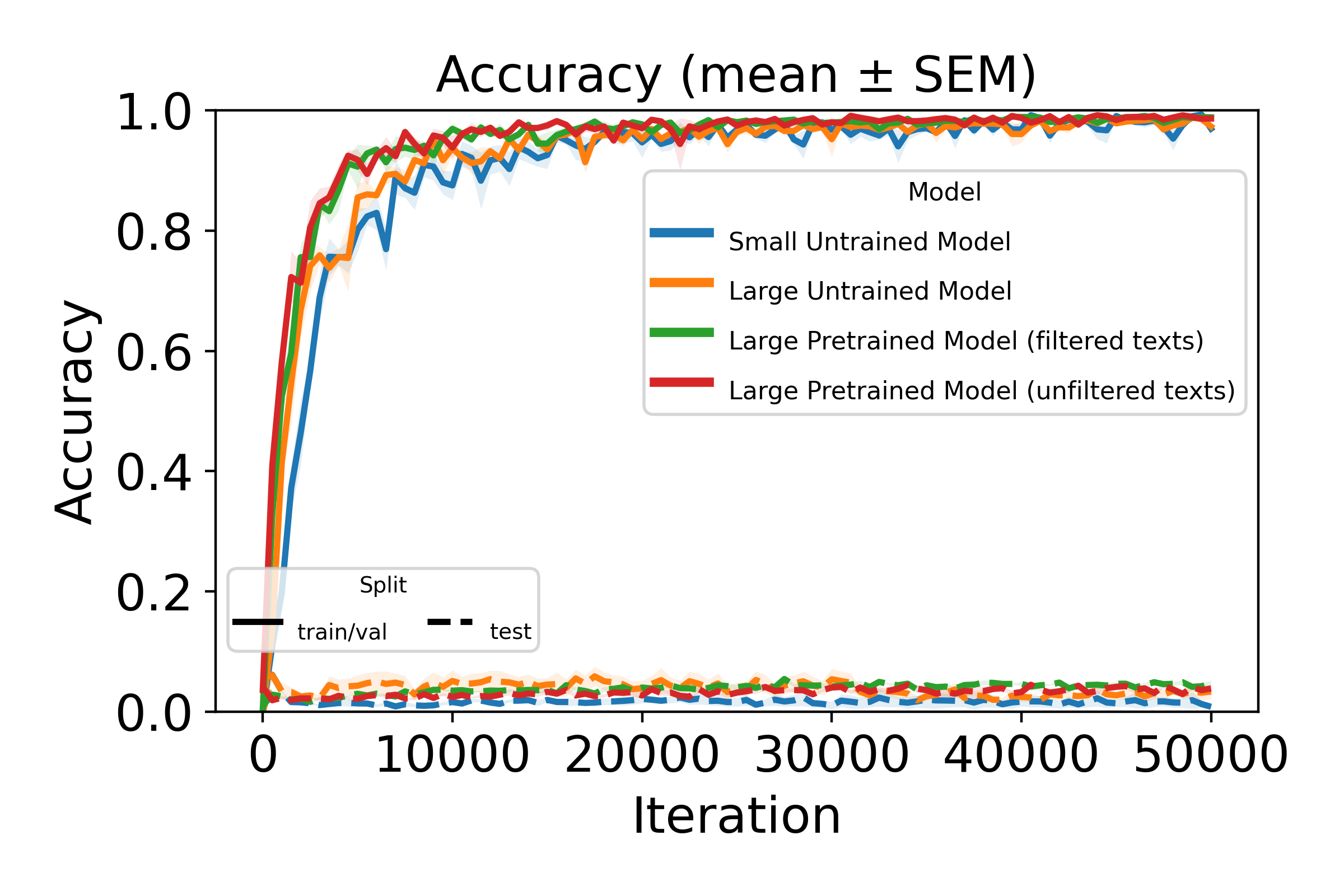}
\end{center}
\noindent \small{Figure 12: Model generalization to zero at test time with answer reversed.}

\begin{center}
\includegraphics[width=0.5\textwidth]{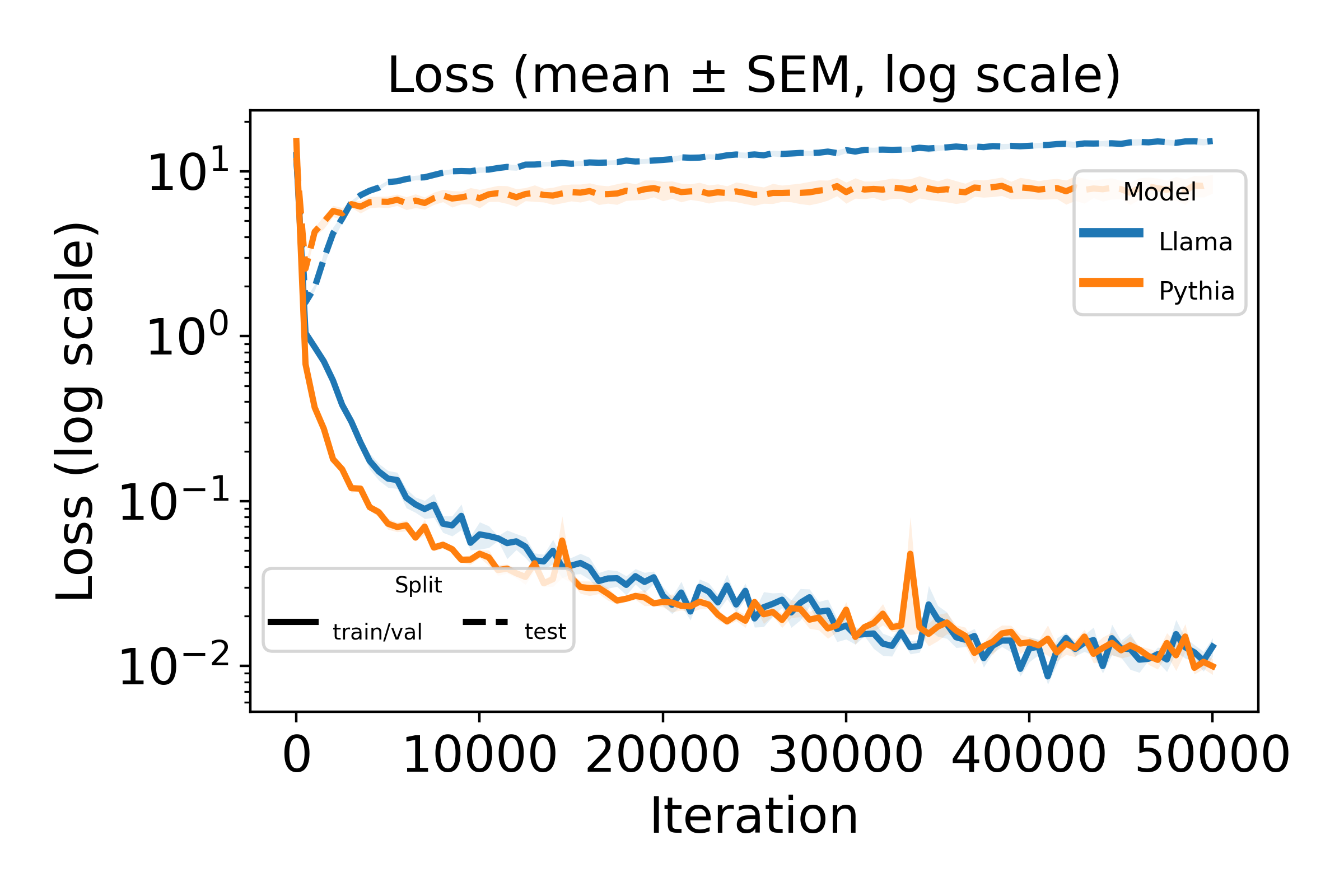}
\end{center}
\noindent \small{Figure 13: (Open sourced) model generalization to zero at test time with answer reversed.}

\begin{center}
\includegraphics[width=0.5\textwidth]{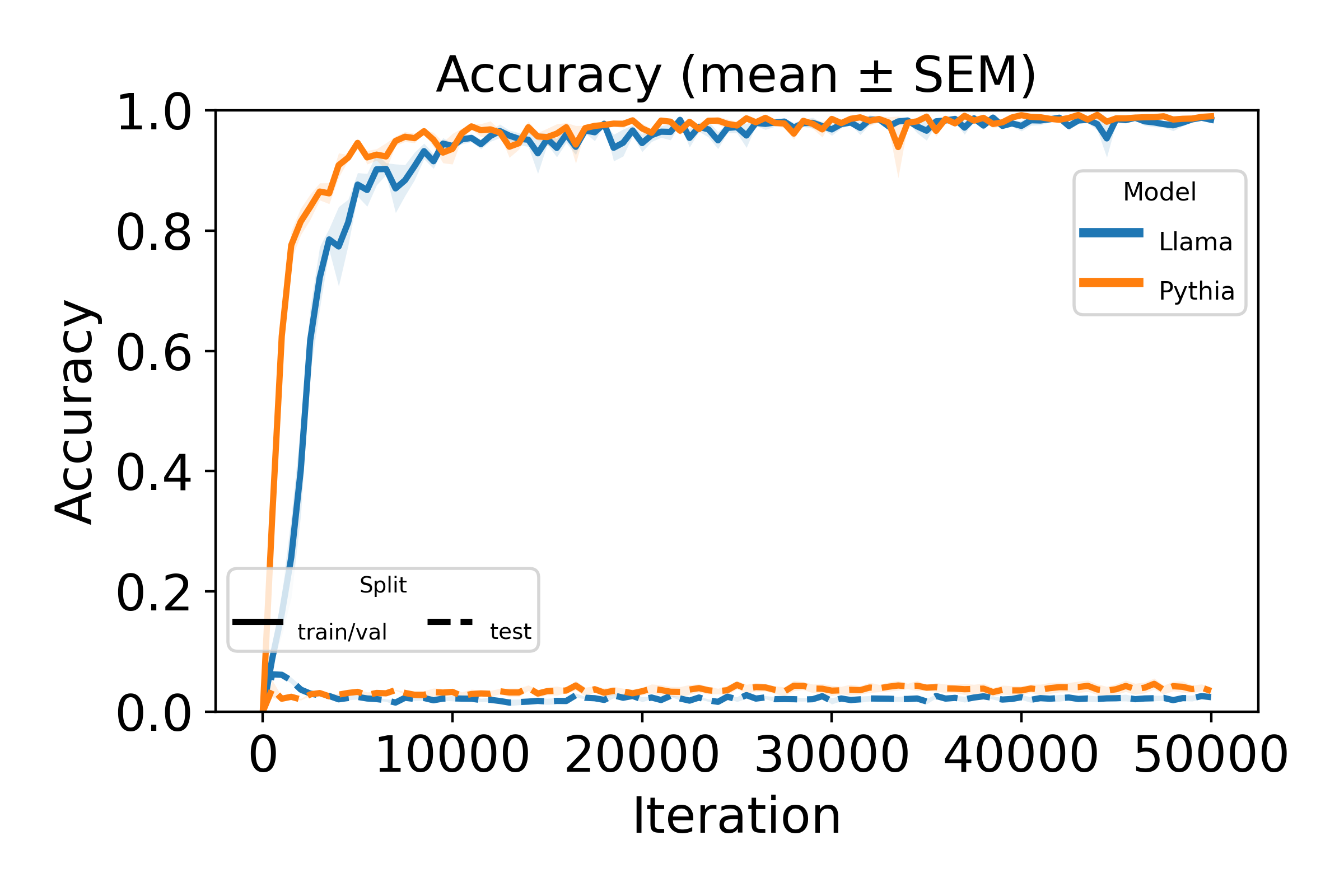}
\end{center}
\noindent \small{Figure 14: (Open sourced) model generalization to zero at test time with answer reversed.}

\subsubsection{Few-shot learning} 

The pretrained model outperforms the untrained model for most few-shot budgets $N$, but the gap in performance is less than when the answer is reversed. Results are visualized in Figure 15.

\begin{center}
\includegraphics[width=0.5\textwidth]{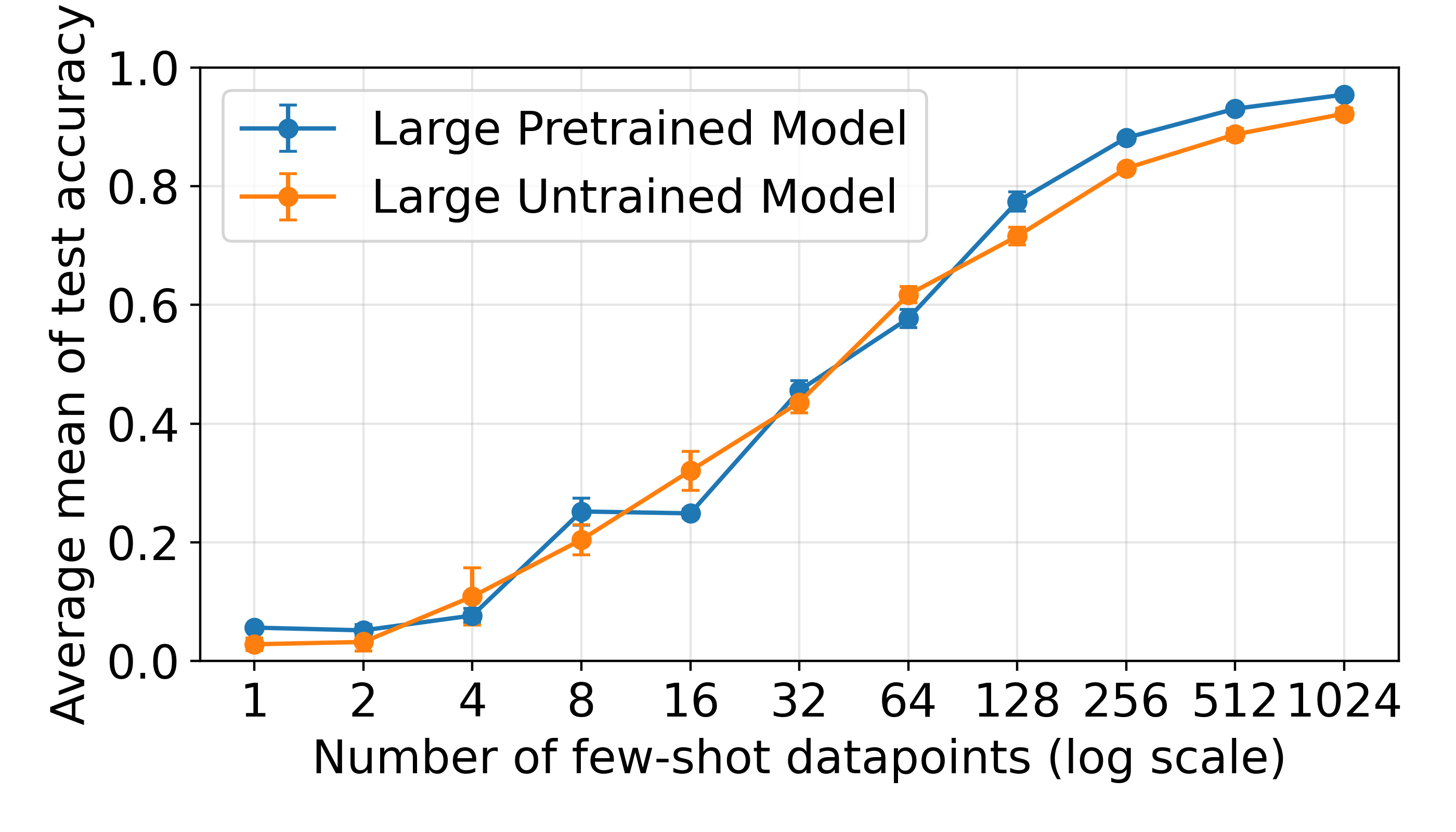}
\end{center}
\noindent \small{Figure 15: Model generalization to zero at test time in few-shot regime with answer reversed.}

\subsubsection{Other numbers} 

The patterns outlined in the main text regarding the other numbers experiments are just as present with answer reversed. In the base-10 case (Figures 16, 17), the digits 0, 9 and 1 are the most difficult to generalize to at test time, followed by 2 and 8, then 7 and 3, and the group least difficult to generalize to is comprised of 4,5 and 6. In the case of base-8 arithmetic (Figures 18, 19), the digits 0,7 and 1 are clearly the most difficult, followed by 2 and 6, generalizing best to the numbers 3,4,5.

\begin{center}
\includegraphics[width=0.5\textwidth]{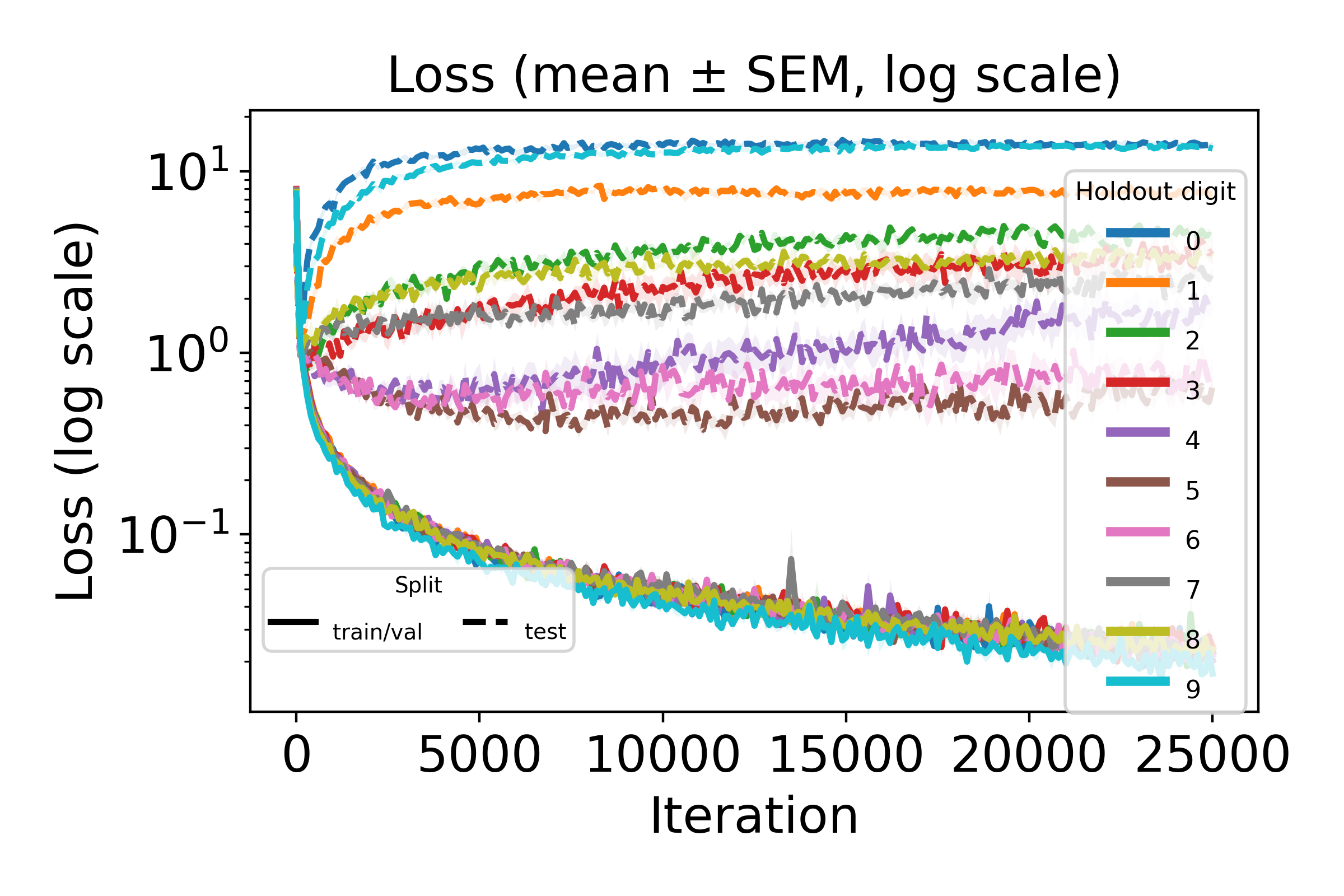}
\end{center}
\noindent \small{Figure 16: Model generalization to digits 0-9 at test time with answer reversed.}

\begin{center}
\includegraphics[width=0.5\textwidth]{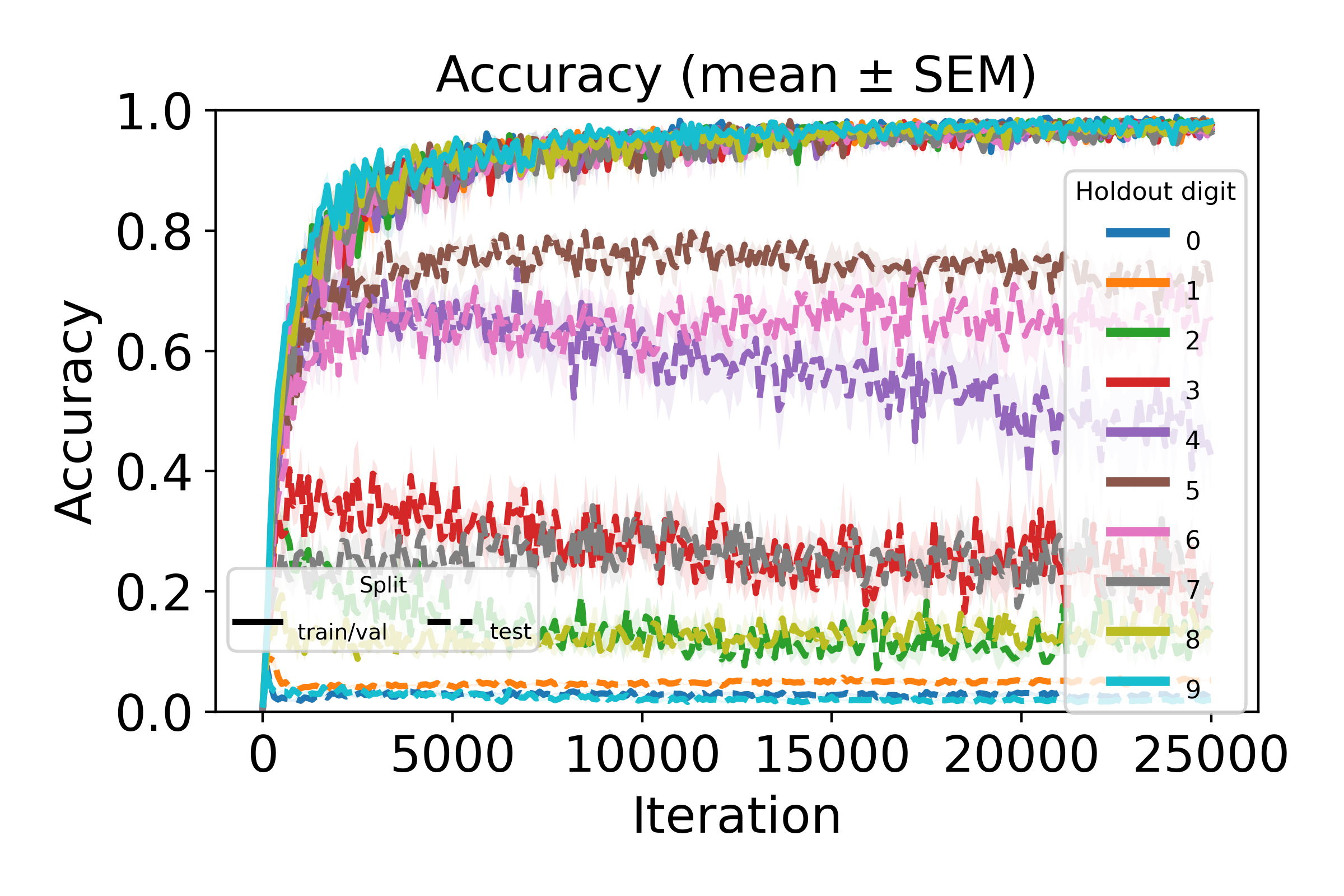}
\end{center}
\noindent \small{Figure 17: Model generalization to digits 0-9 at test time with answer reversed.}

\begin{center}
\includegraphics[width=0.5\textwidth]{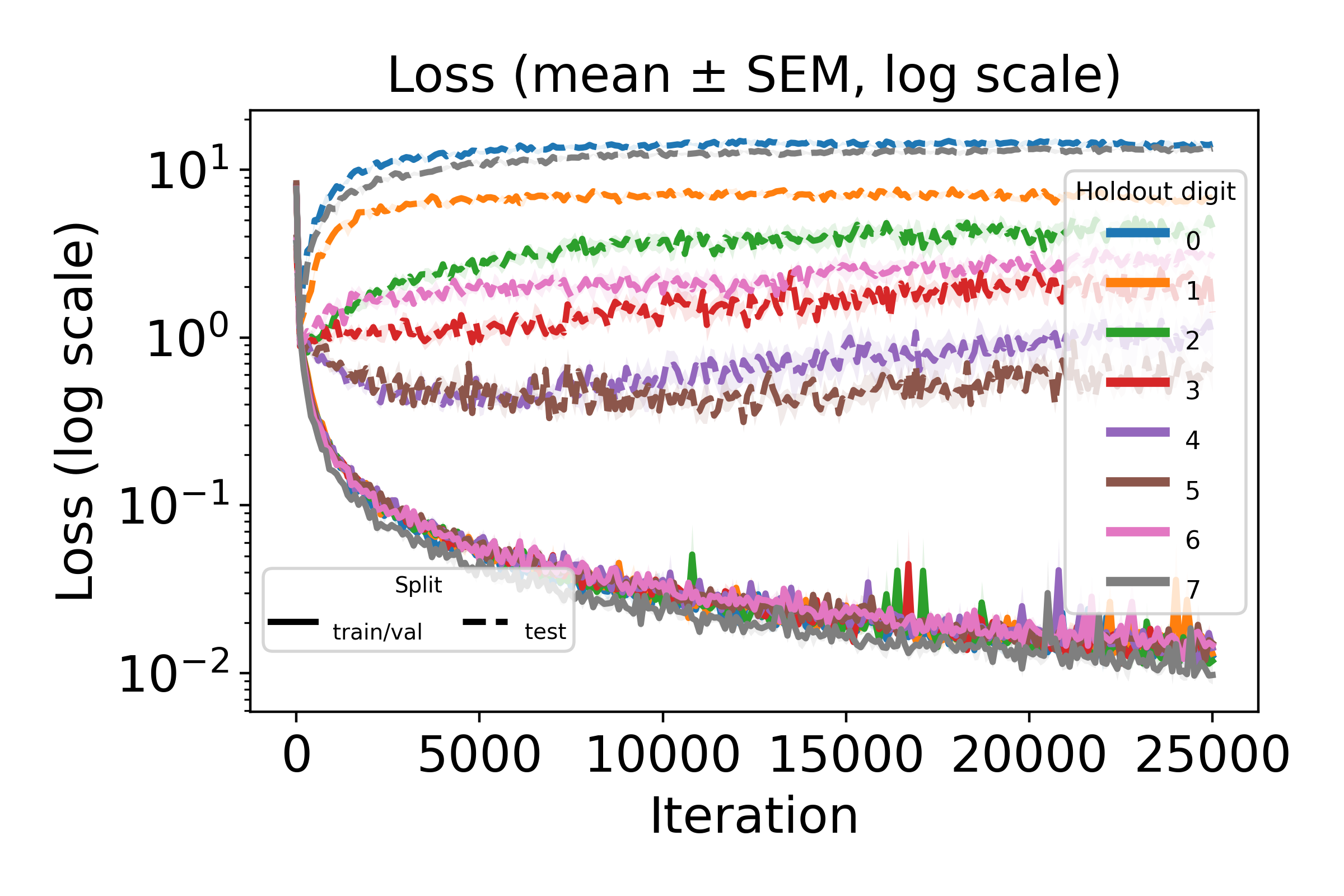}
\end{center}
\noindent \small{Figure 18: Model generalization to digits 0-7 at test time in the base-8 arithmetic regime with answer reversed.}

\begin{center}
\includegraphics[width=0.5\textwidth]{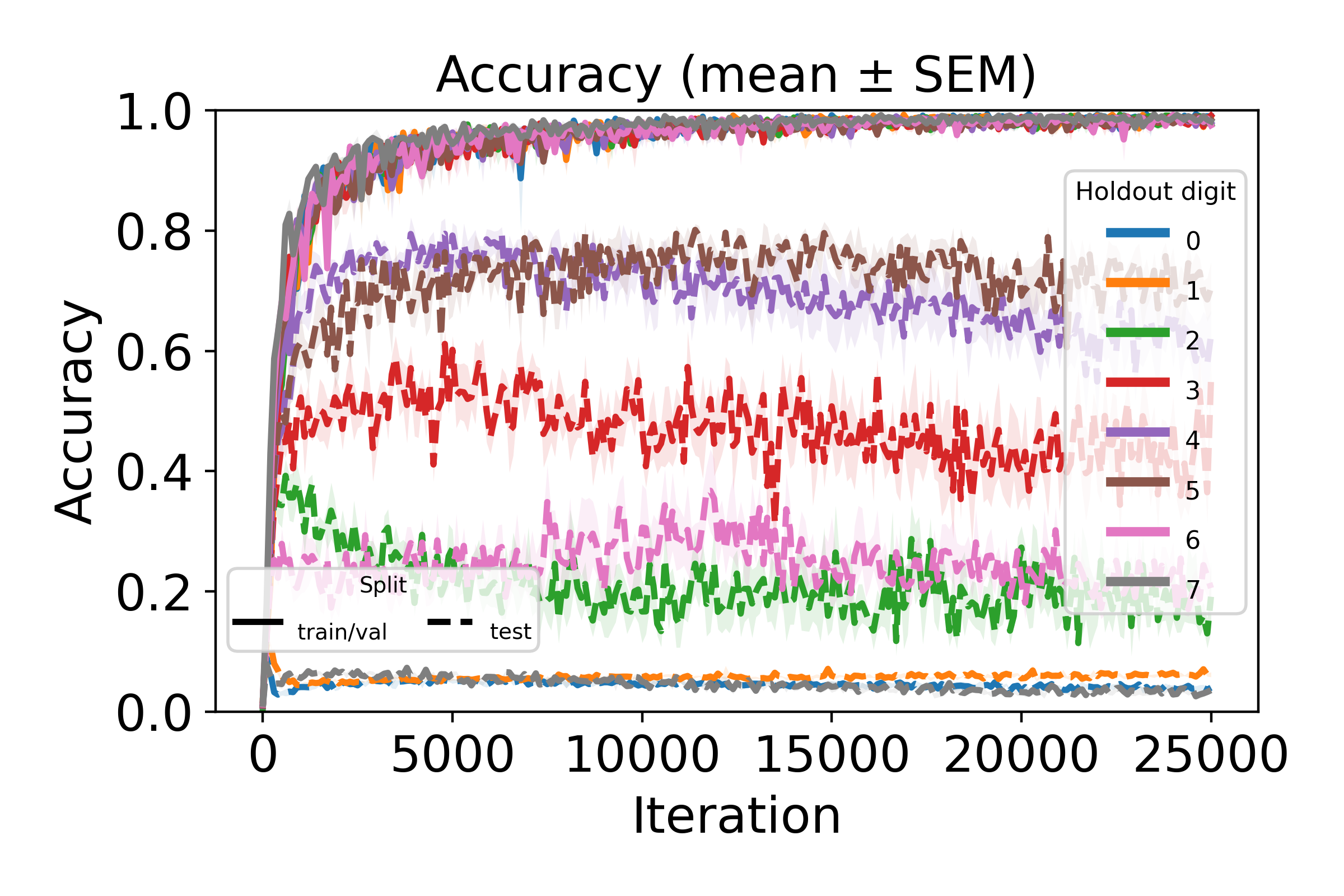}
\end{center}
\noindent \small{Figure 19: Model generalization to digits 0-7 at test time in the base-8 arithmetic regime with answer reversed.}

\subsection{More results for other numbers}

\subsubsection{Training curves}

The patterns outlined in the main text are visible in the full loss and accuracy trajectories of the experiments holding out other numbers. In the base-10 case (Figures 20, 21), three groups of numbers emerge when looking at both test loss and test accuracy. The digits 0, 9 and 1 are the most difficult to generalize to at test time, followed by 2, 3, 7 and 8, the group least difficult to generalize to is comprised of 4,5 and 6. In the case of base-8 arithmetic (Figures 22, 23), this pattern is also present. The digits 0,7 and 1 are clearly the most difficult, followed by 2,3 and 6, generalizing best to the numbers 4,5.

Some of the test-accuracy trajectories increase over the first $\approx 5000$ steps, and then fall. This is best explained by noting that the model often makes errors of a small magnitude on in-distribution examples during the early stages of training. It may be possible that these errors cancel out the errors the model makes at test-time which are due to representing the digit $2$ as $3$ (for example), superficially boosting accuracy. When the model converges on representations for the other digits later in training, the errors due to confusion about the held-out digit remain, resulting in more consistent in-accuracy. Full loss and accuracy trajectories for the other numbers experiments are visualized in Figures 20-23.

\begin{center}
\includegraphics[width=0.5\textwidth]{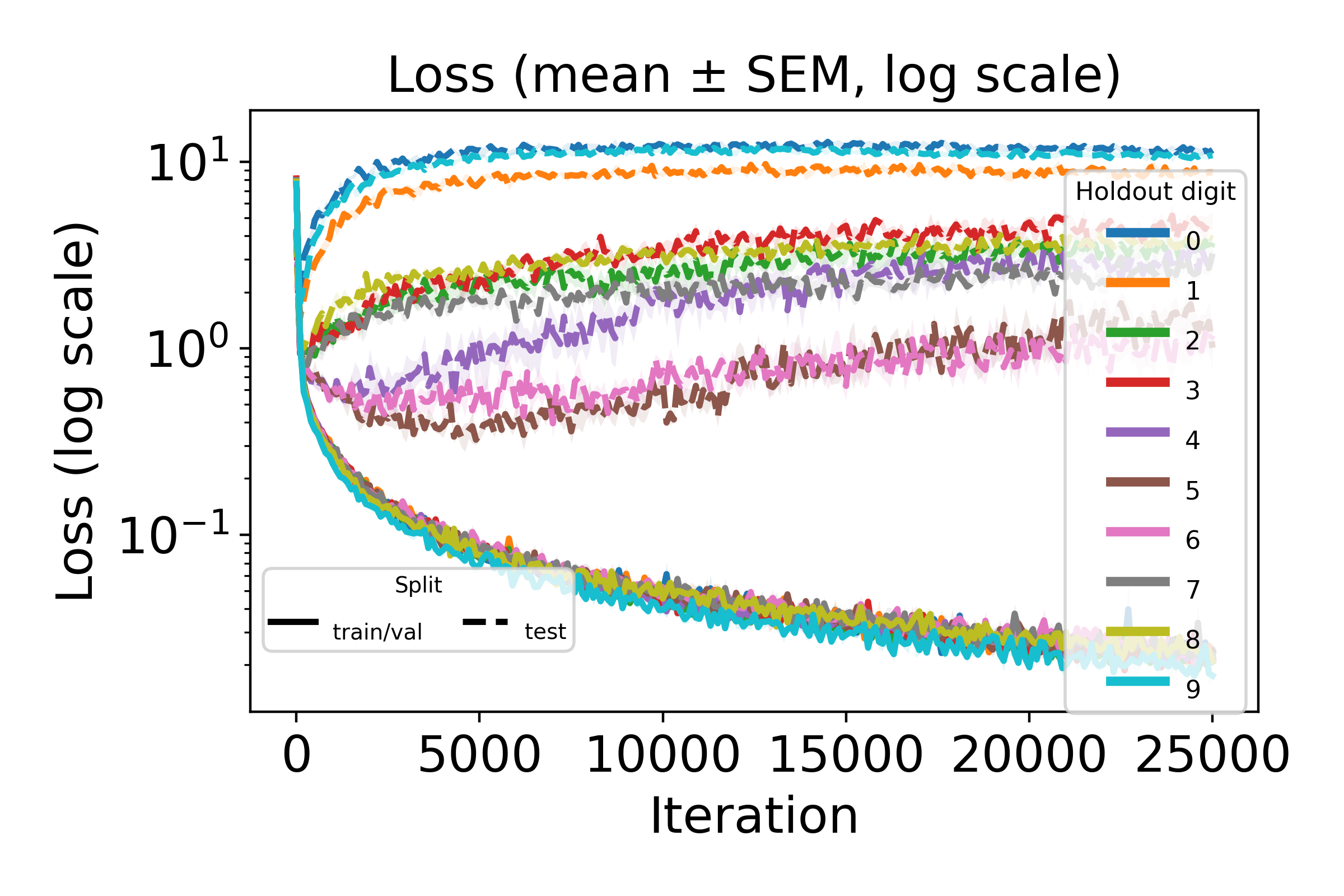}
\end{center}
\noindent \small{Figure 20: Model generalization to digits 0-9 at test time.}

\begin{center}
\includegraphics[width=0.5\textwidth]{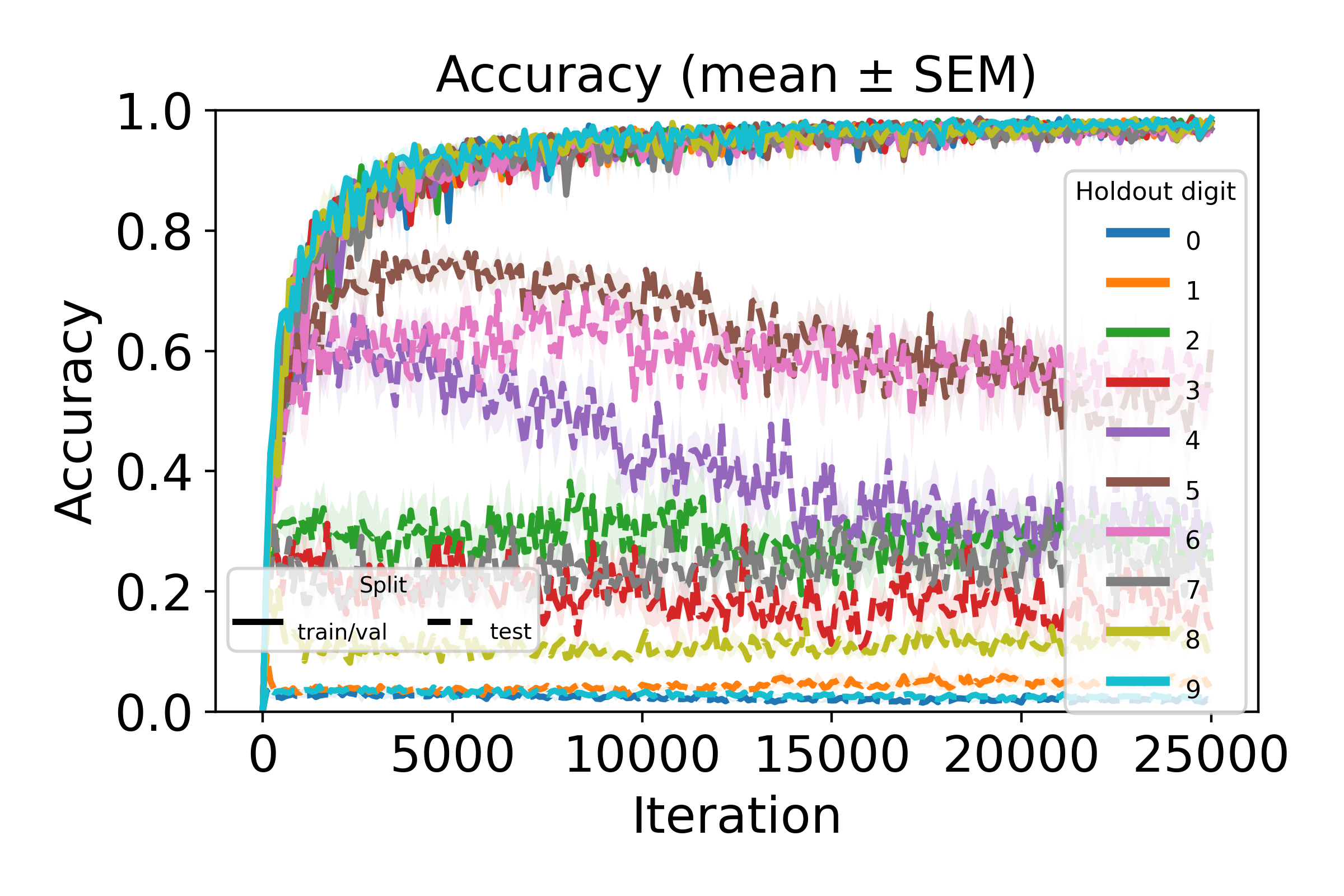}
\end{center}
\noindent \small{Figure 21: Model generalization to digits 0-9 at test time.}

\begin{center}
\includegraphics[width=0.5\textwidth]{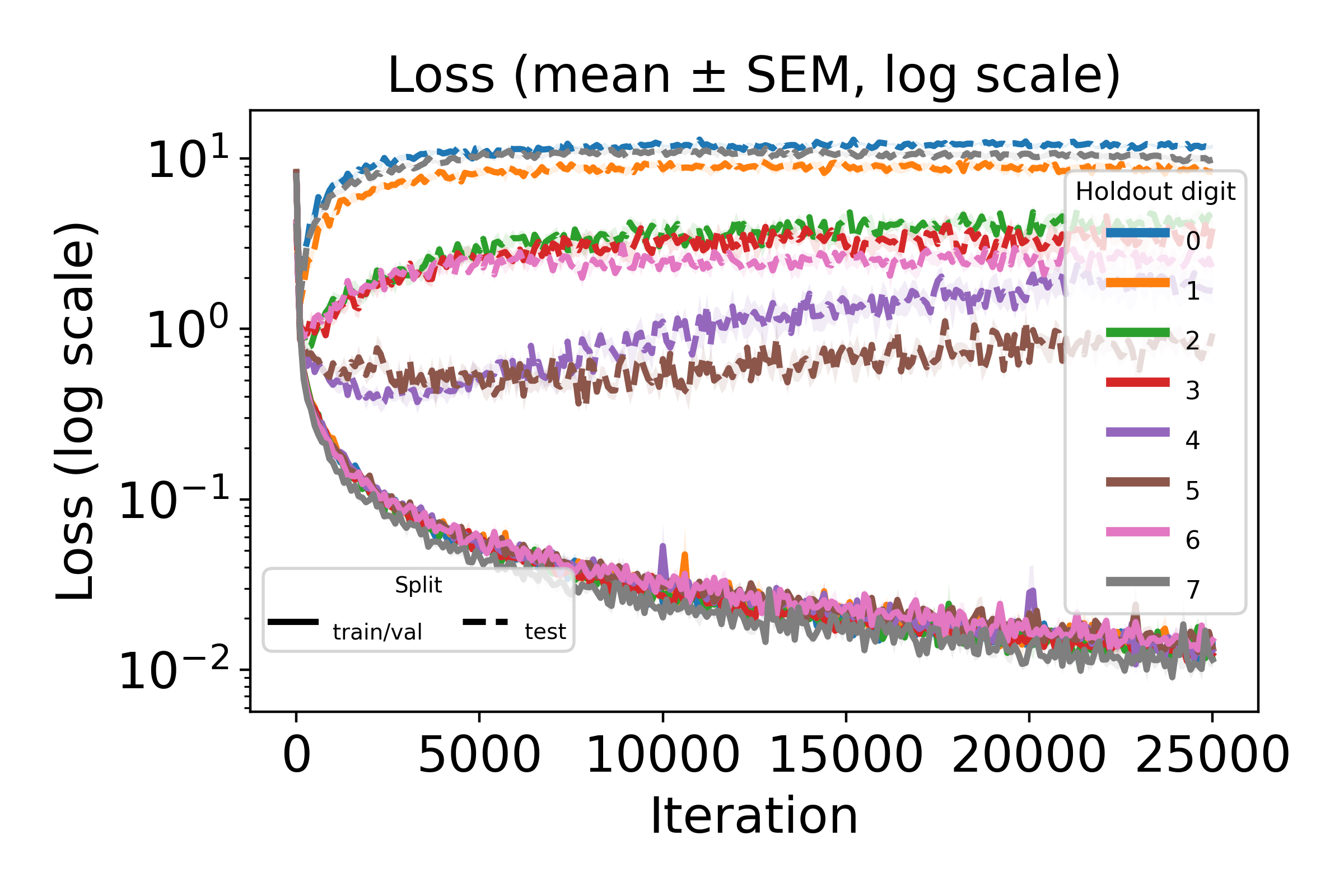}
\end{center}
\noindent \small{Figure 22: Model generalization to digits 0-7 at test time in the base-8 arithmetic regime.}

\begin{center}
\includegraphics[width=0.5\textwidth]{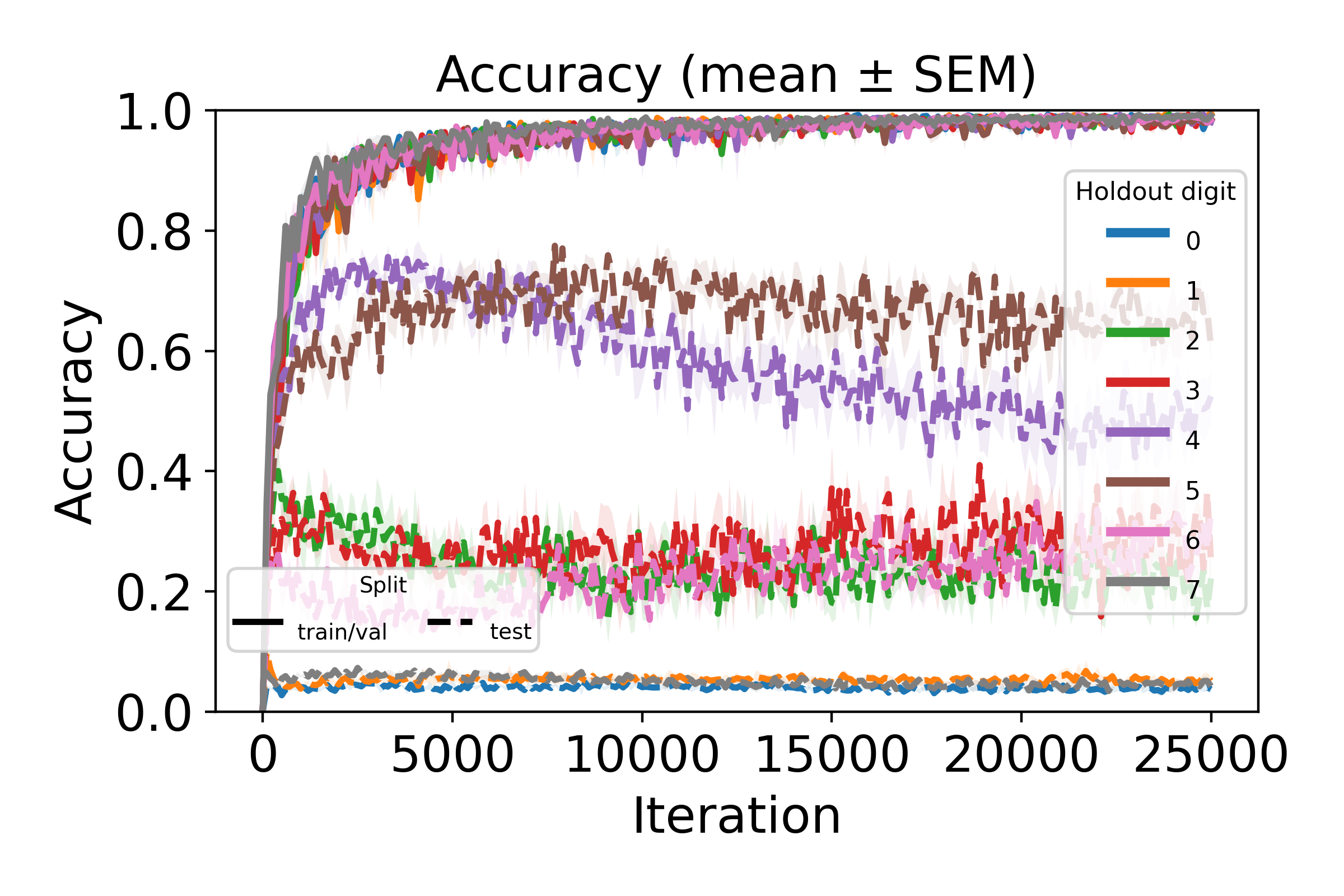}
\end{center}
\noindent \small{Figure 23: Model generalization to digits 0-7 at test time in the base-8 arithmetic regime.}

\subsubsection{Error analysis}

We provide a light analysis of model generalization errors across holdout digits 0-9. Model errors can appear in two places: (Type-1) the model fails to generate the hold-out digit when it is the correct answer to an arithmetic problem (is the RHS), and (Type-2) the model fails to generate the correct answer to an arithmetic problem when the hold-out digit is an operand (on the LHS). 

We report these errors across holdout digits for a single random seed. In particular, for each hold-out digit we report the following: 

\begin{enumerate}
    \item Any surrogate integers that were generated instead of the the hold-out digit (Type-1)
    \item The percent of Type-1 errors that this surrogate integer accounts for 
    \item The percent of Type-2 errors that this surrogate integer accounts for (if we replace all instances of the hold-out digit on the LHS and recompute the answer)
\end{enumerate}

Each row of Table 1 contains these metrics for a unique combination of hold-out digit and surrogate integer. 

\begin{table*}[!ht]
  \begin{center}
    \vskip 0.12in
    \begin{tabular}{rrrr}
      \hline
      Holdout & Surrogate & Percent of RHS Errors (\%) & Percent of LHS Errors (\%) \\
      \hline
      0 & -1 & 59.13 & 0.00 \\
      0 & 1 & 12.17 & 0.00 \\
      0 & -5 & 6.96 & 0.00 \\
      0 & 3 & 5.22 & 15.49 \\
      0 & 4 & 4.35 & 40.71 \\
      0 & 5 & 3.48 & 34.73 \\
      0 & 6 & 3.48 & 2.65 \\
      0 & -4 & 3.48 & 0.00 \\
      0 & 8 & 0.87 & 0.00 \\
      0 & -6 & 0.87 & 0.00 \\
      0 & 2 & 0.00 & 0.11 \\
      1 & 0 & 35.48 & 0.35 \\
      1 & 2 & 33.06 & 21.95 \\
      1 & 3 & 16.13 & 58.89 \\
      1 & -1 & 12.90 & 0.00 \\
      1 & 5 & 1.61 & 1.97 \\
      1 & -2 & 0.81 & 0.00 \\
      1 & 4 & 0.00 & 9.76 \\
      1 & 6 & 0.00 & 0.81 \\
      1 & 7 & 0.00 & 0.35 \\
      1 & 8 & 0.00 & 0.23 \\
      2 & 3 & 71.03 & 76.02 \\
      2 & 1 & 28.28 & 11.13 \\
      2 & 4 & 0.69 & 4.70 \\
      2 & 0 & 0.00 & 0.16 \\
      3 & 4 & 75.83 & 70.16 \\
      3 & 2 & 19.17 & 4.91 \\
      3 & 1 & 4.17 & 0.00 \\
      3 & 5 & 0.83 & 16.84 \\
      4 & 3 & 65.12 & 2.77 \\
      4 & 5 & 34.88 & 78.85 \\
      4 & 6 & 0.00 & 4.55 \\
      5 & 6 & 52.05 & 79.88 \\
      5 & 4 & 47.95 & 8.88 \\
      6 & 5 & 64.66 & 79.24 \\
      6 & 7 & 34.59 & 4.46 \\
      6 & 8 & 0.75 & 0.00 \\
      6 & 4 & 0.00 & 2.68 \\
      7 & 8 & 50.75 & 0.16 \\
      7 & 6 & 49.25 & 88.59 \\
      7 & 5 & 0.00 & 0.64 \\
      8 & 9 & 69.49 & 0.00 \\
      8 & 7 & 28.81 & 88.22 \\
      8 & 6 & 1.69 & 5.89 \\
      9 & 8 & 65.89 & 0.00 \\
      9 & 5 & 15.50 & 36.78 \\
      9 & 6 & 11.63 & 52.98 \\
      9 & 10 & 3.10 & 0.00 \\
      9 & 11 & 3.10 & 0.00 \\
      9 & 3 & 0.78 & 0.00 \\
      9 & 4 & 0.00 & 0.79 \\
      9 & 7 & 0.00 & 2.59 \\
      \hline
    \end{tabular}
    \label{ref:table-1}
  \end{center}
  \caption{Error analysis of Type-1 and Type-2 errors for models trained in the Other Digits experiments.}
\end{table*}

\end{document}